\documentclass[3p,number,fleqn]{elsarticle}

\usepackage[utf8]{inputenc}
\usepackage[OT4]{fontenc}
\usepackage{graphicx}
\usepackage{natbib}
\usepackage{ifpdf}
\ifpdf
    \DeclareGraphicsExtensions{.pdf, .png}
\else
    \DeclareGraphicsExtensions{.eps}
\fi

\usepackage{commath}
\usepackage{mathtools}
\usepackage{epigraph}
\usepackage{enumitem}
\usepackage{textcomp}
\usepackage{amssymb}
\usepackage{amsfonts}
\usepackage{amsthm}
\usepackage{mathrsfs}

\usepackage{stmaryrd}
\usepackage{booktabs}
\usepackage{ifthen}
\usepackage{booktabs}
\usepackage{slashbox}
\usepackage{color}
\usepackage{tabularx}
\usepackage{multirow}
\usepackage{subfig}
\usepackage{caption}
\usepackage{array}
\usepackage[percent]{overpic}
\usepackage[unicode,
					  colorlinks,
						linkcolor={red},
						citecolor={blue},
						urlcolor={blue},
            pdfauthor={Robert Susmaga, Izabela Szczęch, Dariusz Brzezinski},
            bookmarks=false,
            pdfsubject={Towards Explainable TOPSIS: Visual Insights into the Effects of Weights and Aggregations on Rankings},
            pdfkeywords={TOPSIS, weighted criteria ranking, interpretability, visualization, aggregated distance ranking}
            ]{hyperref}

\makeatletter
\newcommand\EqLabel[1]{&\refstepcounter{equation}(\theequation)\ltx@label{#1}&}
\makeatother
\newcolumntype{R}{>{\raggedleft\arraybackslash}X}
\newcolumntype{C}{>{\centering\arraybackslash}X}
\newcolumntype{V}{>{\centering\arraybackslash} m{.45\linewidth} }

\journal{Applied Soft Computing}

\theoremstyle{plain}

\begin{document}
\begin{frontmatter}

\title{Towards Explainable TOPSIS:\\
Visual Insights into the Effects of Weights and Aggregations on Rankings}

\author[1]{Robert Susmaga}
\ead{robert.susmaga@cs.put.poznan.pl}
\author[1]{Izabela Szczęch\corref{cor}}
\ead{izabela.szczech@cs.put.poznan.pl}
\cortext[cor]{Corresponding author}
\author[1]{Dariusz Brzezinski}
\ead{dariusz.brzezinski@cs.put.poznan.pl}

\address[1]{Institute of Computing Science, Poznan University of Technology,\\ Piotrowo 2, 60--965 Poznan, Poland}

\begin{abstract}
Multi-Criteria Decision Analysis (MCDA) is extensively used across diverse industries to assess and rank alternatives. Among numerous MCDA methods developed to solve real-world ranking problems, TOPSIS (Technique for Order Preference by Similarity to Ideal Solution) remains one of the most popular choices in many application areas. TOPSIS calculates distances between the considered alternatives and two predefined ones, namely the ideal and the anti-ideal, and creates a ranking of the alternatives according to a chosen aggregation of these distances. However, interpreting the inner workings of TOPSIS is difficult, especially when the number of criteria is large. To this end, recent research has shown that TOPSIS aggregations can be expressed using the means (M) and standard deviations (SD) of alternatives, creating MSD-space, a tool for visualizing and explaining aggregations. Even though MSD-space is highly useful, it assumes equally weighted criteria, making it less applicable to real-world ranking problems. In this paper, we generalize MSD-space to arbitrary weighted criteria by introducing the concept of WMSD-space defined by what is referred to as weight-scaled means and standard deviations. We demonstrate that TOPSIS and similar distance-based aggregation methods can be successfully illustrated in a plane and interpreted even when the criteria are weighted, regardless of their number. The proposed WMSD-space offers thus a practical method for explaining TOPSIS rankings in real-world decision problems. 
\end{abstract}


\end{frontmatter}
\setlength{\epigraphwidth}{0.42\textwidth}
\epigraph{\textit{Good weight and measure are heaven's treasure.}}{A proverb}

\section{Introduction}
\label{sec:introduction}
Being a sub-discipline of operations research, Multi-Criteria Decision Analysis (MCDA) aims to support decision makers in solving problems that involve real-world objects (alternatives) evaluated on multiple conflicting criteria. Often, this entails selecting the preferred objects, assigning them to preference classes, or ranking them; for an extended overview of MCDA methods, models, and frameworks, see recent reviews \cite{BS_02,Ishizaka13,Bisdorff15,GEF_16,Cinelli22}.
Among methods tackling the task of ranking many alternatives from the most preferred to the least preferred, a commonly chosen approach is TOPSIS (Technique for Order Preference by Similarity to Ideal Solution,~\citet{HWAYOO81}). It is a popular method to create rankings given predefined ideal/anti-ideal alternatives. More precisely, TOPSIS calculates distances from the ideal/anti-ideal to all the alternatives and produces non-negative real values, which determine a linear pre-order that can be used for ranking. 

The TOPSIS method has been widely used in many applications, including logistics~\citep{bottani2006fuzzy}, manufacturing~\citep{wang2009toward,Zhang23ASC}, marketing~\citep{yu2011rank}, sustainable development~\citep{MATEUSZ20181683}, and engineering~\citep{Lin23ASC}; for a much broader survey of TOPSIS and its applications see~\cite{BEHZADIAN201213051,Zavadskas16,Zyoud17}. 
Much of the studies on TOPSIS focus on normalization and weighting procedures. Indeed, criteria weights are an important part of the TOPSIS method, because experts use weights to encode their view of the significance of particular criteria. However, for weights to have the expected effect, criteria values need to first be accordingly normalized. This is particularly important when criteria values are expressed on different scales. The research of \citet{OPRTZE04} has analyzed the impact of different normalization procedures and different aggregation functions on the final ranking obtained using the TOPSIS and VIKOR (VIseKriterijumska Optimizacija i kompromisno Resenje) methods. Similarly, \citet{ZAVZAKANT06} describe the influence of a normalization method on the final TOPSIS rankings. An alternative approach to criterion weighting is considered by \citet{Chakraborty2009ASC}. The topic of weights is also an important part of studies on the Relative Ratio method~\citep{LI09}, which estimates differences between alternatives to create a ranking that balances the distance from the ideal solution and the distance from the anti-ideal solution. Similar approaches to weight balancing and relative closeness have been also proposed by \citet{KUO2017152} and \citet{ABOHADJAM19}. With the use of the ROR (Robust Ordinal Regression) methodology~\cite{Greco2010}, TOPSIS has also been adapted to incorporate predefined relations between alternatives as a form of preferential information from the decision maker~\cite{ZIE17}. Many other interesting issues relating to TOPSIS, including its combinations with other methods, its variations and adaptations, are described in~\cite{CT_23,YYDL-2018,CHEN-2019,TZZL-2018,YK-2017}. 

Last but not least are the attempts aimed at visualizing different dependencies between processed data, and at visualizing the results of ranking methods. 
In the first group, there are typical projection-based approaches, in which new features (attributes) are constructed in order to produce interpretable scatter-plots of some considered objects, e.g., Multi-dimensional Scaling (MDS),~\cite{Cox2008,Borg_05,Walesiak_16} or t-distributed Stochastic Neighbor Embedding (t-SNE),~\cite{t-sne2008,t-sne2012}. These methods are considered interpretable but are a form of lossy compression into at most three new features, making it impossible to reconstruct the original data.  In this sense these methods are imprecise.
Additionally, there are approaches based on considering all the features simultaneously, such as parallel coordinates or radar charts~\cite{Wilke2019}. These are not limited, but their interpretability decreases dramatically as the number of features or alternatives grows. 
Independently, attempts at visualizing the rankings are simply based on one-dimensional graphs depicting the ranking positions of alternatives. 
To the best of our knowledge, apart from the method presented in this paper, there are no methods capable of visualizing the precise ranking and the dependencies in the processed data simultaneously.

As indicated by the referenced works, the bulk of the research on TOPSIS is mainly focused on practical use cases and different ways of performing criteria weighting and normalization. In addition to these application-oriented studies, recently, we have formalized the inner workings of TOPSIS by describing aggregations using the mean (M) and standard deviation (SD) of each alternative. This allowed us to propose a space for visualizing multi-criteria aggregations called MSD-space (Mean-Standard Deviation space) \citep{MSD-space}. Notably, MSD-space is capable of visualizing both data and aggregations, which directly translate to TOPSIS rankings. However, MSD-space assumes equally weighted criteria, making it less applicable to real-world ranking tasks, which, as a recent analysis of TOPSIS applications shows~\cite{BEHZADIAN201213051}, practically always involve decision makers defining weights. By weighing the criteria, they introduce preferential information and alter the influence of the criteria on the final TOPSIS ranking. Oftentimes, a separate MCDA method is applied to elucidate the criteria weights~\cite{lu2023improved,chen2021effects}. That is why applying weights is crucial for modern TOPSIS variations and real-world ranking tasks requiring explainability.

Explainability, regarded as methods that allow humans to understand and trust the results of algorithms, keeps gaining a lot of attention in artificial intelligence \citep{Guidotti_18,Pradhan23ASC,Itani20ASC} and multi-criteria decision support \citep{ZPN_23,Frontiers22,DEBOCK2023}. The ability to explain the result of algorithms is of particular importance in practical applications that impact society. However, such applications often involve domain experts who incorporate their preferential bias in the form of weights imposed on the criteria. Since our existing MSD-space visualizing TOPSIS was designed for unweighted criteria, we pose the following research question: \textit{Is it possible to propose a visualization method that generalizes MSD-space to weighted criteria?}

To answer the above research question, the specific goal of this study is to extend the MSD-space methodology to problems with arbitrarily defined weights of criteria. Moreover, we aim to explain how weights affect rankings of alternatives under various TOPSIS aggregations and how the effects of weights provided by multiple experts can be compared. The detailed contributions of this paper are as follows:
\begin{itemize}
    \item In Section~\ref{sec:preliminaries}, we formalize the TOPSIS procedure and recall the definitions of utility space, MSD-space, and their properties. 
    \item In Section~\ref{sec:weights_and_WMSD}, we put forward our new visualization method. In particular, in Sections~\ref{sec:weights} and \ref{sec:weighted_US} we show how arbitrary criteria weights can be re-scaled and used to generalize utility space into weighted utility space.
    In Section~\ref{sec:IA-WMSD_property}, we define weight-scaled means and standard deviations as equivalents of alternative means and standard deviations in the weighted utility space. As a result, we prove the IA-WMSD property (distance to Ideal and Anti-ideal versus Weighted Mean and  Standard Deviation property) and, in Section~\ref{sec:WMSD-space}, we introduce WMSD-space (Weighted Mean-Standard Deviation space) that represents alternatives in two dimensions regardless of the number of analyzed criteria and their weights. Finally, in Section~\ref{sec:visualization}, we visualize WMSD-space and show how it can be used to express various aggregation functions using the weight-scaled means and standard deviations of alternatives.
    \item In Section ~\ref{sec:case}, we apply the proposed WMSD visualization to two case studies. We show how WMSD-space can be used to explore the properties of a given dataset, compare the effects of weights defined by different experts and underline the implications of using different aggregations.
    \item In Sections~\ref{sec:discussion} and \ref{sec:conclusions}, we discuss our findings and limitations of the proposed method, summarize the paper, and suggest further research.
\end{itemize}

\section{Preliminaries}
\label{sec:preliminaries}
\begin{figure*}[!htb]
\centering
\includegraphics[width=0.95\textwidth]{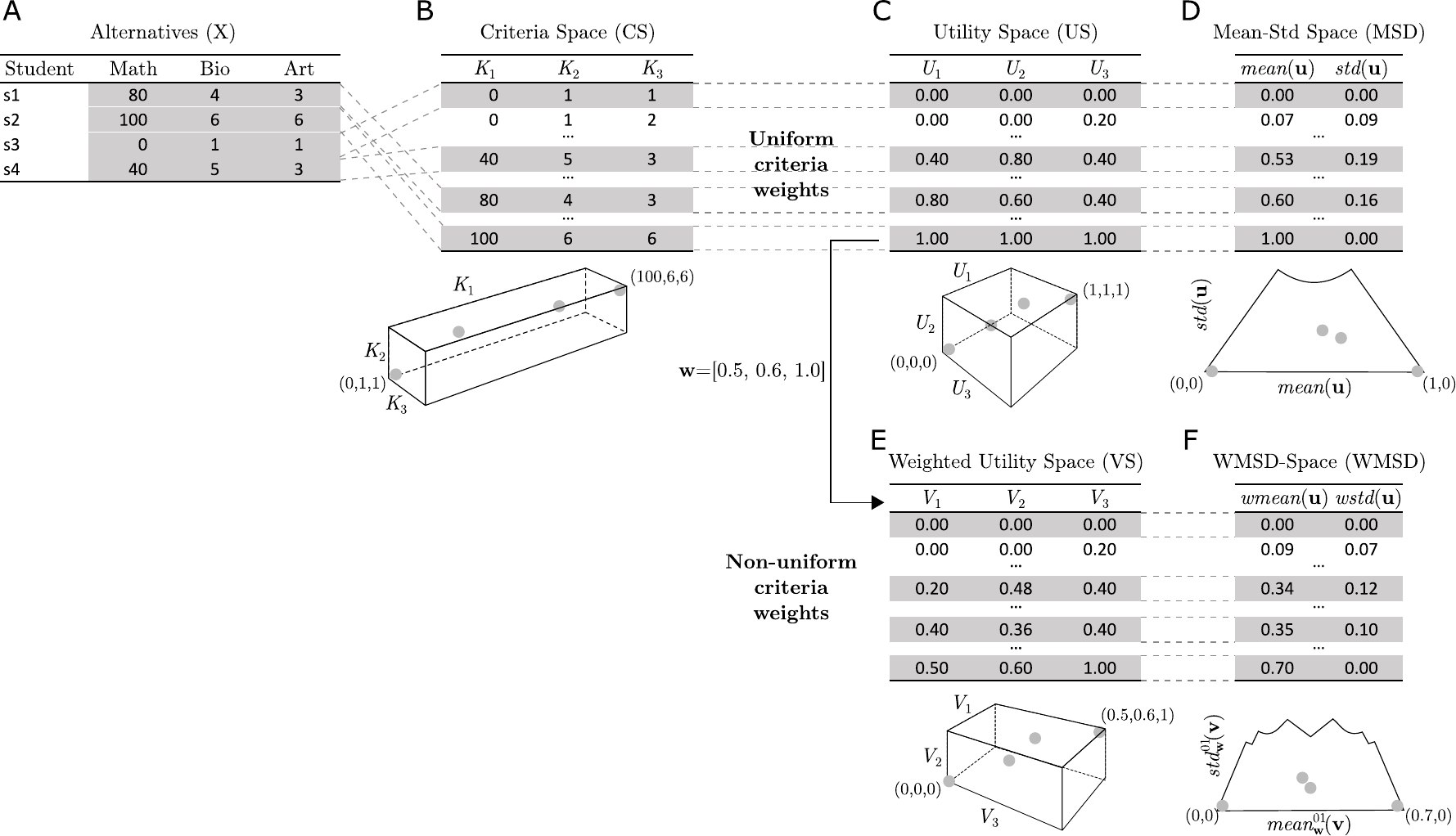}
\caption{The dataset that will serve as the running example for explaining different representations of objects analyzed in this paper. (A) The original dataset (decision matrix) describing $m = 4$ students (alternatives) using final grades from $n = 3$ subjects (criteria). (B) The same dataset depicted as a subset of the criteria space, i.e., of all possible alternatives described by the three criteria describing students. (C) The same alternatives presented as a subset of utility space, the re-scaled equivalent of criteria space. (D) The analyzed students represented in MSD-space, a space defined by the mean (M) and standard deviation (SD) of the utility space descriptions of the alternatives. (E) Alternatives represented in weighted utility space, with weights $\mathbf{w} = [0.5, 0.6, 1.0]$. (F) Alternatives represented in WMSD-space, a space defined by the weight-scaled mean (WM) and weight-scaled standard deviation (WSD) of the weighted utility space descriptions of the alternatives.}
\label{fig:all-spaces}
\end{figure*}

The majority of research on the Technique for Order of Preference by Similarity to Ideal Solution (TOPSIS) method involves a predefined, finite set of $m$ entities (referred to as \textit{alternatives}) described by a set of $n$ features (\textit{criteria}). Consequently, the information can be effectively represented in an $m \times n$ matrix of values, commonly known as the \textit{decision matrix}. An example of such a decision matrix $\mathbf{X}$ is illustrated in Figure~\ref{fig:all-spaces}A, comprising $m = 4$ alternatives (students) characterized by $n = 3$ criteria (final grades in subjects).

Unlike research papers that focus on specific applications of TOPSIS, our study will not be confined to a specific set of $m$ alternatives. Instead, we will delve into the general characteristics of all possible alternatives given a set of $n$ criteria. Our analysis of all conceivable alternative representations is influenced by strategies designed for visually inspecting general properties of machine learning metrics~\citep{DBLP:journals/isci/BrzezinskiSSS18,TetrahedronDemo,AMCS2015,BullPass2015}. To conduct this analysis, in this section, we will provide the essential definitions needed to formalize the TOPSIS procedure. Furthermore, we will review the conclusions from our previous research on interpreting TOPSIS by revisiting the definitions of utility space, IA-MSD property (distance to Ideal and Anti-ideal versus Mean and Standard Deviation Property), and MSD-space (Mean-Standard Deviation space)~\citep{MSD-space}. The notation introduced in the following paragraphs will be used to generalize utility space and MSD-space into their weighted counterparts in the subsequent sections of this paper.

\subsection{Formalizing TOPSIS Using the Utility Space}
\label{sec:Formalizing_TOPSIS}
\noindent
TOPSIS \citep{HWAYOO81} is a multi-criteria decision analysis (MCDA) method that ranks objects (\textit{alternatives}) from the best to the worst in terms of their distance to ideal and anti-ideal points. 
The descriptions of alternatives with respect to considered attributes is commonly given in the form of vectors. Among attributes typically used in MCDA, there are \textit{criteria}, characterized by preference-ordered domains. 

The main actions performed by TOPSIS method can be summarized as: 
\begin{enumerate}
\item \textbf{Prepare the representations} of alternatives in terms of criteria. Apart from forming the decision matrix, this part of the procedure may also normalize the criteria and incorporate the user-given weights that actually constitute their preferential information.
\item \textbf{Determine two reference points}, ideal and anti-ideal, and verify how far each alternative is from them.  
\item \textbf{Rank the alternatives} with respect to some aggregation function that combines the distances between the alternatives and the ideal/anti-ideal points.
\end{enumerate}

\paragraph{Preparing representations of alternatives} TOPSIS starts with encoding real-world objects (e.g., students described by criteria referring to their grades) into a decision matrix $\mathbf{X}$ (Figure~\ref{fig:all-spaces}A). The decision matrix is a finite subset of the criteria space $\mathit{CS}$ (Figure~\ref{fig:all-spaces}B), where if a criterion $\mathcal{K}$ belongs to the set of all possible criteria $\mathbb{K}$ ($\mathcal{K} \in \mathbb{K}$), then its domain is a real-valued interval $\mathcal{V} = [v_{\mathit{min}},v_{\mathit{max}}]$. 
Since TOPSIS is based on calculating distances, the bounds of the interval need to be finite.
Additionally, criteria, as attributes with preferentially ordered domains, may differ in their preference types (gain or cost), with the least preferred value denoted as $v_*$ and the most preferred value as $v^*$. 
Vectors $[v_1^*, v_2^*, ..., v_n^*]$ and $[v_{1*}, v_{2*}, ..., v_{n*}]$ will be referred to as the ideal ($I$) and anti-ideal ($A$) points, respectively.

Working on criteria with varying domains and types can make the analysis more troublesome and reduce the meaningfulness of the results, thus a criteria transformation is often applied. In this paper, we will use a min-max re-scaling that transforms the criteria space into the utility space $\mathit{US}$ (Figure~\ref{fig:all-spaces}C) using the function $\mathcal{U}: \mathcal{V} \rightarrow [0,1]$:
\begin{itemize}
\item given a domain $\mathcal{V} = [v_{\mathit{min}},v_{\mathit{max}}] = [v_*,v^*]$ of a criterion $\mathcal{K} \in \mathbb{K}$ of type gain, the re-scaling function $\mathcal{U}$ associated with $\mathcal{K}$ is defined as $\mathcal{U}(v) = \frac{v-v_*}{v^*-v_*}$ for $v \in \mathcal{V}$,
\item given a domain $\mathcal{V} = [v_{\mathit{min}},v_{\mathit{max}}] = [v^*,v_*]$ of a criterion $\mathcal{K} \in \mathbb{K}$ of type cost, the re-scaling function $\mathcal{U}$ associated with $\mathcal{K}$ is defined as $\mathcal{U}(v) = \frac{v_* - v}{v_*-v^*}$ for $v \in \mathcal{V}$.
\end{itemize}

The $\mathcal{U}(\cdot)$ function is introduced to simplify further TOPSIS processing without the loss of generality and is independent of decision matrix normalization that could be performed by a user. 
The motivations behind introducing the min-max transformation are two-fold.  First, the counter-domain of the transformation is the [0,1] interval, which combined with the value reversal for cost type criteria produces well interpretable results: the most preferred values are always transformed to ones, while the least preferred values are always transformed to zeros. Second, after proper selection of the min and max values for each criterion, application of the min-max transformation liberates TOPSIS from the peril of the so-called rank-reversal problem, to which original TOPSIS is unfortunately predisposed~\citep{MARROY06}. 

Since the $\mathit{US}$ is the space of all conceivable representations (images) of alternatives, particular decision matrices are simply represented as finite subsets of $\mathit{US}$. Aiming at formalizing general dataset-independent properties, we shall deploy $\mathit{US}$ in all further considerations.

\paragraph{Determination of the ideal/anti-ideal points and distance calculation}
Given a set of criteria $\mathscr{K}$, $|\mathscr{K}| = n \geq 1$, the utility space is an $n$-dimensional hypercube $[0,1] \times [0,1] \times \cdots \times [0,1]$ with $2^n$ vertices of the form $[z_1, z_2, ..., z_n]$, where $z_j \in \{0, 1\}$. Moreover, for each alternative representation $\mathit{E} \in \mathit{CS}$ there exists $\mathbf{u} \in \mathit{US}$ such that $\mathbf{u}$ is the image of $\mathit{E}$ under the re-scaling transformation---if $\mathit{E} = [v_1, v_2, ..., v_n] \in \mathit{CS}$, then $[\mathcal{U}_1(v_1), \mathcal{U}_2(v_2), ..., \mathcal{U}_n(v_n)] \in \mathit{US}$. In particular, $\mathit{US}$ contains vectors $\mathbf{1} = [1, 1 ..., 1]$ and $\mathbf{0} = [0, 0, ..., 0]$, which are the respective images of the ideal point and anti-ideal point. 
In our running example, there are three criteria, thus the points $\mathbf{1} = [1, 1, 1]$ and $\mathbf{0} = [0, 0, 0]$ represent in $\mathit{US}$ the ideal point and anti-ideal point, respectively (see Figure~\ref{fig:all-spaces}C).

With the $\mathbf{1}$ and $\mathbf{0}$ points at hand, TOPSIS calculates how far each alternative is from them.
To perform this operation, the Euclidean distance measure is used.

Let $\mathbf{a} = [a_1,a_2,...,a_n]$ and $\mathbf{b} = [b_1,b_2,...,b_n]$ be vectors (always assumed to be row vectors). The dot product of these vectors is defined as $\mathbf{a} \cdot \mathbf{b}^T = [a_1,a_2,...,a_n] \cdot [b_1,b_2,...,b_n]^T = \sum_{j=1}^{n}a_j \cdot b_j$ (the second vector is transposed to allow matrix multiplication). This serves to define the Euclidean norm of vector $\mathbf{a} = [a_1,a_2,...,a_n]$ as its dot product with itself: $\norm{\mathbf{a}}_2 = \sqrt{\mathbf{a} \cdot \mathbf{a}^T} = \sqrt{\sum_{j=1}^{n}a_j^2}$. Finally, the Euclidean distance between vectors $\mathbf{a} = [a_1,a_2,...,a_n]$ and $\mathbf{b} = [b_1,b_2,...,b_n]$ is defined as the Euclidean norm of their difference: $\delta_2(\mathbf{a},\mathbf{b}) = \norm{\mathbf{a} - \mathbf{b}}_2 = \sqrt{\sum_{j=1}^{n}(a_j-b_j)^2}$. Now, the Euclidean distance is used to compute $\delta_2(\mathbf{u},\mathbf{0})$ and $\delta_2(\mathbf{u},\mathbf{1})$ for each $\mathbf{u} \in \mathit{US}$.

The maximal Euclidean distance in $\mathit{US}$, which extends between vectors $\mathbf{1}$ and $\mathbf{0}$, is dependent on $n$ and equals $\sqrt{n}$. 
For our analyses to be $n$-independent and thus easily interpretable regardless of $n$, we define a re-scaled Euclidean distance as $\delta^{01}_2(\mathbf{a},\mathbf{b}) = \frac{\delta_2(\mathbf{a},\mathbf{b})}{\sqrt{n}}$, ranging always between $[0,1]$ (instead of $[0, \sqrt{n}]$).
The re-scaled distances of an alternative's image $\mathbf{u} \in \mathit{US}$ to the ideal and anti-ideal point will be denoted as $\delta^{01}_2(\mathbf{u},\mathbf{1})$ and $\delta^{01}_2(\mathbf{u},\mathbf{0})$, respectively.

\paragraph{Ranking alternatives according to an aggregation}
The distances of each alternative's representation to the reference points are combined with respect to some chosen aggregation function, the value of which naturally forms a ranking of the alternatives (precisely: a linear pre-order).

We shall focus on three aggregations, defined in terms of $\delta^{01}_2(\mathbf{u},\mathbf{1})$ (`distance to the ideal in $\mathit{US}$') and $\delta^{01}_2(\mathbf{u},\mathbf{0})$ (`distance to the anti-ideal in $\mathit{US}$') as:
\begin{align} 
\mathsf{I}(\mathbf{u}) &= 1 - \delta^{01}_2(\mathbf{u},\mathbf{1}), \\
\mathsf{A}(\mathbf{u}) &= \delta^{01}_2(\mathbf{u},\mathbf{0}), \\
\mathsf{R}(\mathbf{u}) &= \frac{\delta^{01}_2(\mathbf{u},\mathbf{0})}{\delta^{01}_2(\mathbf{u},\mathbf{1})+\delta^{01}_2(\mathbf{u},\mathbf{0})},
\end{align}
where $\mathbf{u} \in \mathit{US}$ is the image of an alternative in $\mathit{US}$. All three aggregations generate values belonging to $[0, 1]$. The $\mathsf{I}(\mathbf{u})$ aggregation is based solely on the distance to the \textit{ideal} point, the $\mathsf{A}(\mathbf{u})$ aggregation is based on the distance to the \textit{anti}-ideal point, whereas the \textit{relative} distance (the standard aggregation of TOPSIS), denoted as $\mathsf{R}(\mathbf{u})$, takes both previous distances into account.
Using $1 - \delta^{01}_2(\mathbf{u},\mathbf{1})$ instead of a straightforward distance $\delta^{01}_2(\mathbf{u},\mathbf{1})$ in the $\mathsf{I}(\mathbf{u})$ aggregation,
serves only as a means to have all aggregations as functions to be maximized.
Although only the $\mathsf{R}(\mathbf{u})$ aggregation is predominantly deployed in TOPSIS, it is defined on the basis of $\mathsf{I}(\mathbf{u})$ and $\mathsf{A}(\mathbf{u})$, and as such inherits from them its main properties. For this reason, all three aggregations will be examined in this paper.

\subsection{The IA-MSD Property and MSD-space}
\begin{figure*}[!h]
\centering
\includegraphics[width=\textwidth]  {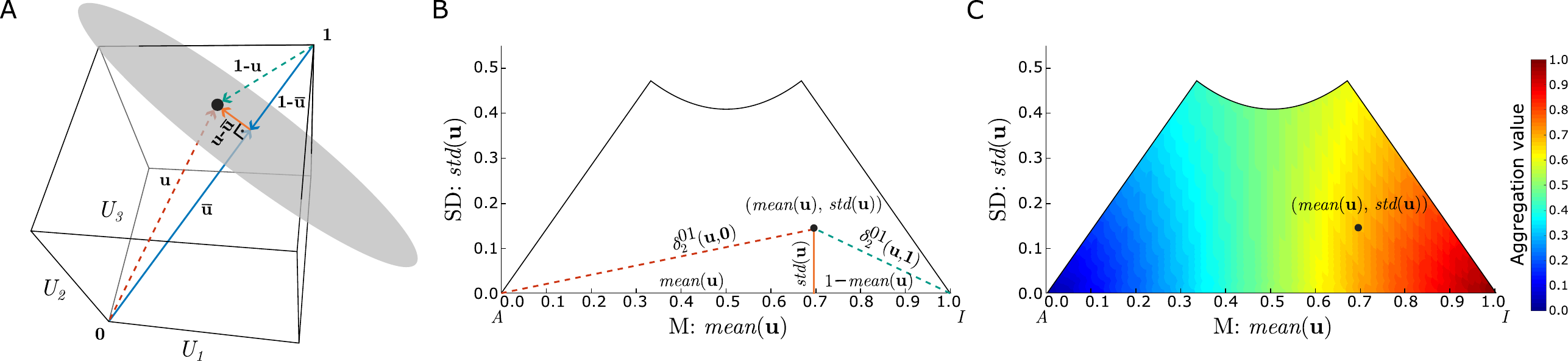}
\caption{A depiction of the IA-MSD property in $\mathit{US}$ and MSD-space for a three-dimensional problem. (A) Vector orthogonality depicted in $\mathit{US}$. (B) Illustration of the IA-MSD property in MSD-space. The re-scaled $\delta^{01}_2$ lengths of vectors $\mathbf{\overline{u}}$ and $\mathbf{u} - \mathbf{\overline{u}}$ from panel A correspond to the values of $mean(\mathbf{u})$ and $std(\mathbf{u})$ depicted in MSD-space. (C) Color encoding of the aggregation function $\mathsf{R}(\mathbf{u})$, with blue representing the least preferred and red the most preferred values.}
\label{fig:MSD_tutorial}
\end{figure*}

Given a representation of an alternative in utility space $\mathbf{u} \in \mathit{US}$, let:
\begin{align}
sum(\mathbf{u}) &= \sum_{j=1}^nu_j,\\
mean(\mathbf{u}) &= \frac{sum(\mathbf{u})}{n},\\
var(\mathbf{u}) &= \frac{\norm{\mathbf{u}-\mathbf{\overline{u}}}_2^2}{n}, \, \textrm{with} \,\,\, \mathbf{\overline{u}} = [mean(\mathbf{u}), mean(\mathbf{u}), ..., mean(\mathbf{u})],\\
std(\mathbf{u}) &= \sqrt{var(\mathbf{u})}.
\end{align}
With the above notation, in our previous paper~\citep{MSD-space}, we have used the fact that 
for every $\mathbf{u} \in \mathit{US}$ vectors 
$\mathbf{\overline{u}}-\mathbf{0}$ and $\mathbf{u} - \mathbf{\overline{u}}$ 
as well as
$\mathbf{u} - \mathbf{\overline{u}}$ and $\mathbf{1} - \mathbf{\overline{u}}$ 
are orthogonal, and, therefore, one can apply the Pythagorean theorem to relate these vectors (Figure~\ref{fig:MSD_tutorial}A). 

Moreover, in~\citep{MSD-space} we have shown that the lengths of the above-mentioned vectors can be expressed as follows:
\begin{align} 
    \delta^{01}_2(\mathbf{\overline{u}},\mathbf{0}) &= mean(\mathbf{u}),\\
    \delta^{01}_2(\mathbf{\overline{u}},\mathbf{1}) &= 1-mean(\mathbf{u}),\\
    \delta^{01}_2(\mathbf{u},\mathbf{\overline{u}}) &= std(\mathbf{u}).
\end{align}

These characteristics of $\mathit{US}$ allowed us to formulate the \textit{IA-MSD property}.

\theoremstyle{definition}
\newtheorem{definition}{Definition}
\newtheorem{theorem}{Theorem}

\begin{theorem}[IA-MSD Property]
\hfill
\begin{align}
\delta^{01}_2(\mathbf{u},\mathbf{0}) &= 
\sqrt{mean(\mathbf{u})^2+std(\mathbf{u})^2}, \\
\delta^{01}_2(\mathbf{u},\mathbf{1}) &= 
\sqrt{(1-mean(\mathbf{u}))^2+std(\mathbf{u})^2}.
\end{align}
\end{theorem}

The IA-MSD property shows that the distances of an alternative to the ideal and anti-ideal point are functions of the mean and standard deviation of the alternative. This interesting dependency between the distances of alternatives to the predefined ideal ($I$) and anti-ideal ($A$) points on the one hand and 
$mean(\mathbf{u})$ and $std(\mathbf{u})$ on the other, inspired us to define  \textit{MSD-space}. This space uses the mean (M) and standard deviation (SD) of an alternative's $\mathit{US}$ representation as its constituents (Figure~\ref{fig:MSD_tutorial}B).

\begin{definition}[MSD-space]
\hfill
\begin{align}
\text{MSD-space} =\{[mean(\mathbf{u}),std(\mathbf{u})] \; | \; \mathbf{u} \in US\}
\end{align}
\end{definition}

The MSD-space is a two-dimensional space and, therefore, can be visualized in a plane where the mean (M) of an alternative defines its position on the x-axis and the standard deviation (SD) the position on the y-axis. Since MSD-space is a transformation\footnote{M is a linear function of elements of $\mathbf{u} \in \mathit{US}$, while SD is a square root of a second order polynomial of elements of $\mathbf{u} \in \mathit{US}$.} of $\mathit{US}$ which is $[0,1]$-bounded, the range of values of M and SD is also bounded. As a result, for a given number of criteria $n$, there is a limited range of attainable means and standard deviations, which forms the shape of MSD-space (Figure~\ref{fig:MSD_tutorial}B). 

Moreover, the IA-MSD property makes it possible to define all TOPSIS aggregations in terms of $mean(\mathbf{u})$ and $std(\mathbf{u})$:
\begin{align}
    \mathsf{I}(\mathbf{u})
    &= 1 - \sqrt{(1-mean(\mathbf{u}))^2+std(\mathbf{u})^2}, \\
    \mathsf{A}(\mathbf{u})
    &= \sqrt{mean(\mathbf{u})^2+std(\mathbf{u})^2},\\
    \mathsf{R}(\mathbf{u})
    &= \frac{\sqrt{mean(\mathbf{u})^2+std(\mathbf{u})^2}}{
    {\textstyle
        \sqrt{(1-mean(\mathbf{u}))^2+std(\mathbf{u})^2}
    }
    {\textstyle
        + \sqrt{mean(\mathbf{u})^2+std(\mathbf{u})^2}
    }}
    .
\end{align}
Since all the discussed TOPSIS aggregations are functions of two parameters, $mean(\mathbf{u})$ and $std(\mathbf{u})$, one can visualize their values in MSD-space using a color map (Figure~\ref{fig:MSD_tutorial}C). Using such a visualization, one can analyze how the preferences expressed by different aggregations change with varying values of M and SD.

\section{WMSD-space: a method for visualizing TOPSIS aggregations and rankings}
\label{sec:weights_and_WMSD}
The concepts of utility space, IA-MSD property, and MSD-space proposed in~\citep{MSD-space}, and recalled in the previous section, assume that all of the criteria are equally important. In practice, this is rarely the case, as criteria are very often differentiated and thus assigned non-uniform weights by experts. In this section, we will formalize criteria weighting and show how the utility space $\mathit{US}$ can be transformed to its weighted counterpart $\mathit{VS}$. We will then define the weight-scaled mean (WM) and weight-scaled standard deviation (WSD), which will be used to introduce a weighted version of the MSD-space, called the WMSD-space. To make it easier to follow and compare the introduced concepts, Table~\ref{tab:notation} summarizes the mathematical notation used to define MSD-space in the previous section and WMSD-space presented in this section.

\renewcommand{\arraystretch}{1.2}
\begin{table}[!h]
    \centering
    \caption{Notation summary.}
    \footnotesize
    \begin{tabularx}{\textwidth}{lll}
    \toprule
        & Symbol & Definition \\
    \midrule
		\multirow{19}{*}{MSD}& $\mathit{CS}$ & Criteria space.\\
        & $\mathbf{X} \subset \mathit{CS}$ & Decision matrix. \\
		& $E \in \mathit{CS}$ & Alternative. \\
		& $n$ & Number of criteria. \\
		& $m$ & Number of alternatives in $\mathbf{X}$. \\
		& $\mathbb{K}$ & Set of criteria.\\
		& $\mathcal{K} \in \mathbb{K}$ & Criterion. \\
        & $v$ & Criterion value. \\
		& $\mathcal{V} = [v_{\mathit{min}},v_{\mathit{max}}]$ & Domain of a criterion. \\
        & $v_*$; $v^*$ & Least preferred; most preferred criterion value. \\
		& $\mathcal{U}: \mathcal{V} \rightarrow [0,1]$ & Rescaling function that transforms the criteria space into the utility space. \\
		& $\mathit{US}$ & Utility space.\\
        & $D_{\mathbf{0}}^{\mathbf{1}}$ & Diagonal of $\mathit{US}$. \\
		& $\mathbf{u} \in \mathit{US}$ & Representation (image) of an alternative in utility space (normalized alternative). \\
  	& $\mathbf{\overline{u}}$ & Vector of means ($[mean(\mathbf{u}), ..., mean(\mathbf{u})]$). \\
        & $|\cdot|$ & Set cardinality.\\
        & $\norm{\,\cdot\,}_2$ & Euclidean norm. \\ 
		& $\delta_2(a, b)$ & Euclidean distance. \\
		& $\delta^{01}_2(a, b)$ & 0-1 re-scaled Euclidean distance. \\
		& $mean(\mathbf{u})$, M & Mean of normalized alternative. \\
		& $std(\mathbf{u})$, SD & Standard deviation of normalized alternative. \\
		& $\mathsf{I}(\mathbf{u}), \mathsf{A}(\mathbf{u}), \mathsf{R}(\mathbf{u})$ & Ideal ($\mathsf{I}$); anti-ideal ($\mathsf{A}$); and relative distance ($\mathsf{R}$) aggregation for \textit{unweighted} criteria. \\
  
	\midrule

        \multirow{9}{*}{WMSD} & $\mathbf{w}$ & Criteria weights. \\
        & $\mathit{VS}$ & Weighted utility space. \\
        & $D_{\mathbf{0}}^{\mathbf{w}}$ & Diagonal of $\mathit{VS}$. \\
        & $\mathbf{v} \in \mathit{VS}$ & Representation (image) of an alternative in weighted utility space (weighted alternative). \\
		& $\mathbf{\overline{v}}$ & Vector of weighted means ($\mathbf{w} \circ \mathbf{\overline{u}}$). \\
        & $n_p$ & Number of non-zero weighted criteria. \\
        & $s = \frac{\norm{\mathbf{w}}_2}{mean(\mathbf{w})}$ & Weight-scaling factor. \\
        & $\delta^{01}_{\mathbf{w}}(\mathbf{a},\mathbf{b}) = \frac{\delta_2(\mathbf{a},\mathbf{b})}{s}$ & Weight-scaled Euclidean distance.\\
        & $mean^{01}_{\mathbf{w}}(\mathbf{v})$, WM & Weight-scaled mean of a weighted alternative.\\
        & $std^{01}_{\mathbf{w}}(\mathbf{v})$, WSD & Weight-scaled standard deviation of a weighted alternative.\\
        & $\mathsf{I}_{\mathbf{w}}(\mathbf{v}); \mathsf{A}_{\mathbf{w}}(\mathbf{v}); \mathsf{R}_{\mathbf{w}}(\mathbf{v})$ & Ideal ($\mathsf{I}$); anti-ideal ($\mathsf{A}$); relative distance ($\mathsf{R}$) aggregation for \textit{weighted} criteria. \\
    \bottomrule
    \end{tabularx}
    \label{tab:notation}
\end{table}
\renewcommand{\arraystretch}{1}

\subsection{Normalized Criteria Weights}
\label{sec:weights}
Let $\mathbf{w} = [w_1, w_2, ..., w_n]$ be a vector of real values, acting as criteria weights (referred to as weights), where $n$ is the number of criteria. According to common MCDA practice, these weights are assumed to be:
\begin{itemize}
\item all non-negative,
\item at least one non-zero,
\item all finite.
\end{itemize}
These assumptions have the following justifications.

Non-negativity results from the fact that the weight expresses the magnitude of the criterion's relative importance (greater weight, greater importance). In particular, if $w_i > w_j$, then the $i$-th criterion is expected to have more influence on the final result of the method than the $j$-th criterion. 
Therefore, the first considered assumption is: $w_i \geq 0$ for all $i$. 

Next, notice that relation `$\geq$' admits two disjoint sub-cases: `$=$' and `$>$'. Whenever $w_i = 0$, then the $i$-th criterion is in practice `zeroed' and thus not taken into account in any further considerations. On the other hand, $w_i \neq 0$ means that the $i$-th criterion is taken into account. 
This explains the second assumption, namely: $w_i \neq 0$ for at least one $i$ (`at least one non-zero'). The assumption ensures that the undesired case of all zero weights cannot occur. If all weights were zero, the descriptions of all alternatives would be identical and there would be no need to subject the alternatives to any ranking method.

Finally, we shall consider bounded, and thus finite, weights (`all finite'). 
Such a constraint may be implemented by `normalizing' the weights to satisfy $\max_{i=1}^{n}w_i = 1$ (`max is $1$'). This is achieved by dividing all of the weights by their maximum, which is positive thanks to the two previous assumptions. Such an operation does not affect the `performance' of weights, because their relative ratios remain the same after any division by a positive number.

As a result, the three assumptions (`all non-negative', `at least one non-zero', and `all finite'), are expressed with only two conditions:
\begin{align}
&\mathbf{w} \geq \mathbf{0}\ \text{(`all non-negative'),}\\
&\max_{i=1}^{n}w_i = 1\ \text{(`max is $1$', ensuring `at least one non-zero' and `all finite').}
\end{align}

It should be stressed that even though the postulated assumptions exclude situations in which \emph{all} weights are zero, they do not exclude situations in which \emph{some} weights are zero. As stated above, in such a situation, the criteria corresponding to zero weights are in practice eliminated from all further considerations. In result, `zeroing' weights may be viewed as a form of `criterion selection' (only criteria corresponding to positive weights are selected).

In the following sections, we will assume that criteria weights adhere to the `all non-negative' and `max is 1' conditions. This can be easily implemented in practice, as any set of real values satisfying `all non-negative', `at least one non-zero' and `all finite' can be re-scaled to be `all non-negative' and `max is 1'.

\subsection{$\mathit{VS}$: The Weighted Utility Space}
\label{sec:weighted_US}
While $\mathit{US}$ has the shape of a hypercube, it changes as soon as it becomes non-uniformly weighted, i.e., as soon as non-uniform  criteria weights are applied. By weighing the criteria, one introduces preferential information from the decision maker and alters the influence of the criteria on the final TOPSIS ranking (as will be later illustrated in case studies in Section~\ref{sec:case}). Thus, the shape of the weighted version of $\mathit{US}$, in which all TOPSIS operations are actually performed, generalizes to a hyperrectangle, with the special case of the hypercube obtained for all weights equal to one. Notice that the application of weights is in fact a linear operation.

\begin{figure*}[h!]
\centering
\includegraphics[width=0.75\textwidth]  {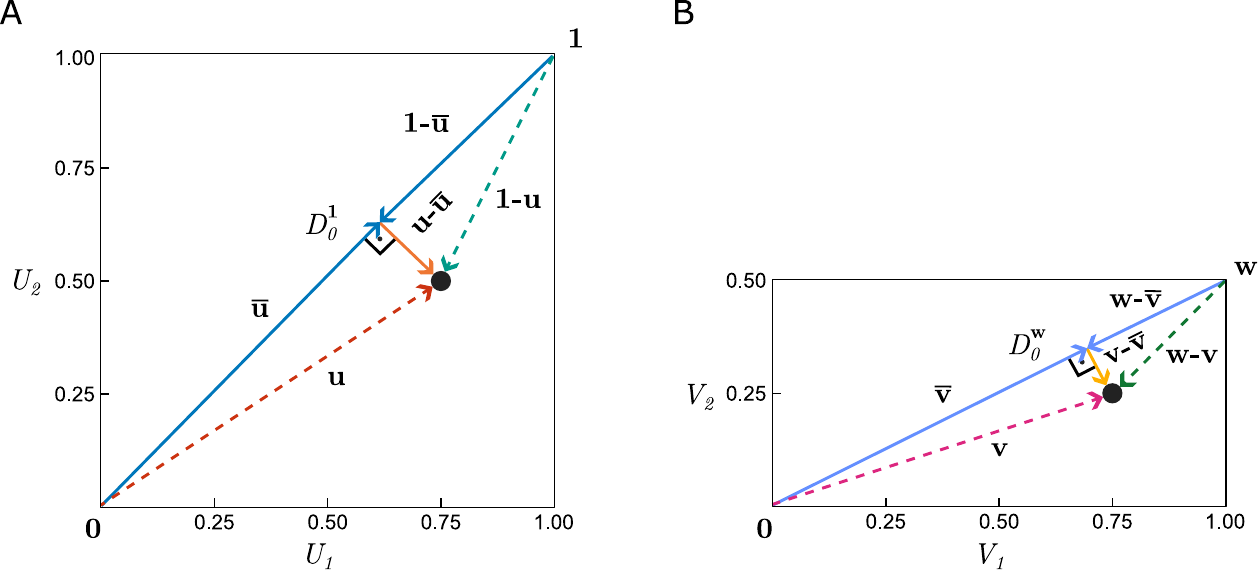}
\caption{Vector orthogonality presented in (A) $\mathit{US}$ and (B) $\mathit{VS}$, for $n = n_p = 2$. The weight vector used to transform the presented $\mathit{US}$ into $\mathit{VS}$ is $\mathbf{w} = [1.0, 0.5]$. The diagonal $D_{\mathbf{0}}^{\mathbf{1}}$ is the blue line segment between vertices $\mathbf{0}$ and $\mathbf{1}$  in $\mathit{US}$. Analogously, $D_{\mathbf{0}}^{\mathbf{w}}$ is the blue line segment between vertices $\mathbf{0}$ and $\mathbf{w}$ in $\mathit{VS}$.}
\label{fig:Orthogonality_2D}
\end{figure*}

Given vectors $\mathbf{a} = [a_1, a_2, ..., a_n]$ and $\mathbf{b} = [b_1, b_2, ..., b_n]$, let $\mathbf{a} \circ \mathbf{b}$ denote their element-wise (Hadamard) product, i.e., $\mathbf{a} \circ \mathbf{b} = [a_1 \cdot b_1, a_2 \cdot b_2, ..., a_n \cdot b_n]$. Now, let $\mathbf{w} = [w_1, w_2, ..., w_n]$ be a vector of  weights. Given these weights and an $n$-dimensional $\mathit{US}$ we define 
\begin{equation}
  \mathit{VS} = \{\mathbf{v}: \mathbf{v} = \mathbf{w} \circ \mathbf{u}, \mathbf{u} \in US\}. 
\end{equation}
The weighted utility space $\mathit{VS}$ (Figure~\ref{fig:Orthogonality_2D}) is the image of $\mathit{US}$ with (in particular):
\begin{itemize}
\item $\mathbf{0} \in \mathit{VS}$ being the image of $\mathbf{0} \in \mathit{US}$,
\item $\mathbf{w} \in \mathit{VS}$ being the image of $\mathbf{1} \in \mathit{US}$.
\end{itemize}
\noindent 
By the assumptions of $\mathbf{w}$ (`max is $1$'), if $\mathbf{u} \in \mathit{US}$, then $\mathbf{v} = \mathbf{w} \circ \mathbf{u} \leq \mathbf{u}$. In result, $\mathbf{1} \not \in \mathit{VS}$ in general (this only happens when $\mathbf{w} = \mathbf{1}$, since then $\mathit{VS} = \mathit{US}$ and $\mathbf{1} \in \mathit{VS}$ is the image of $\mathbf{1} \in \mathit{US}$).
Clearly, $\mathit{VS} \subseteq \mathit{US}$, with $\mathit{VS} = \mathit{US}$ only for $\mathbf{w} = \mathbf{1}$; in all other cases $\mathit{VS} \subset \mathit{US}$. Additionally, while $\mathit{US}$ is an $n$-dimensional hypercube, $\mathit{VS}$ is represented by a $n_p$-dimensional hyperrectangle, where 
\begin{equation}
 n_p = |\{i: w_i > 0\}|. 
\end{equation}
The assumption `at least one non-zero' of $\mathbf{w}$ ensures that $n_p \geq 1$, so in general, $1 \leq n_p \leq n$. In the most often satisfied case of $w_i > 0$ for all $i$, $n_p = n$. 

Similarly to $\mathit{US}$, two vertices, namely $\mathbf{w}$ and $\mathbf{0}$ (images of $\mathbf{1}$ and $\mathbf{0}$ from $\mathit{US}$), are of special interest in $\mathit{VS}$, as they constitute the endpoints of the segment that will be referred to as the main diagonal of $\mathit{VS}$ and denoted as $D_{\mathbf{0}}^{\mathbf{w}}$ (Figure~\ref{fig:Orthogonality_2D}). The diagonal $D_{\mathbf{0}}^{\mathbf{w}}$ is an image of $D_{\mathbf{0}}^{\mathbf{1}} = \{\mathbf{d} = d \cdot \mathbf{1} \; | \; d \in [0, 1]\}$ (the diagonal of $\mathit{US}$), since it contains all vectors $\{\mathbf{w} \circ \mathbf{d} \; | \; \mathbf{d} \in D_{\mathbf{0}}^{\mathbf{1}}\}$, in particular $\mathbf{w}$ (for $\mathbf{d} = \mathbf{1}$) and $\mathbf{0}$ (for $\mathbf{d} = \mathbf{0}$). Of course $D_{\mathbf{0}}^{\mathbf{w}}$ thus equals $\{d \cdot \mathbf{w} \; | \; d \in [0, 1]\}$.

As was the case with $D_{\mathbf{0}}^{\mathbf{1}} \subseteq \mathit{US}$, $D_{\mathbf{0}}^{\mathbf{w}}$ satisfies $D_{\mathbf{0}}^{\mathbf{w}} \subseteq \mathit{VS}$, but is dependent on $n_p$ (rather than $n$), as $D_{\mathbf{0}}^{\mathbf{w}} \subset \mathit{VS}$ for $n_p > 1$ and $D_{\mathbf{0}}^{\mathbf{w}} = \mathit{VS}$ for $n_p=1$.
 
The maximal Euclidean distance $\delta_2$ in $\mathit{VS}$ is that of $D_{\mathbf{0}}^{\mathbf{w}}$, which extends between vectors $\mathbf{0}$ and  $\mathbf{w}$ (Figure~\ref{fig:Orthogonality_2D}). This maximal distance equals $\norm{\mathbf{w}}_2$, which makes it heavily dependent on $\mathbf{w}$.
To make the maximal distance in $\mathit{VS}$ independent of at least some characteristics of $\mathbf{w}$, we define the re-scaled weighted Euclidean distance $\delta^{01}_{\mathbf{w}}$. Given any $\mathbf{v_{1}}, \mathbf{v_{2}} \in \mathit{VS}$:
\begin{equation}
\delta^{01}_{\mathbf{w}}(\mathbf{v_{1}},\mathbf{v_{2}}) = \frac{\delta_2(\mathbf{v_{1}},\mathbf{v_{2}})}{s}, 
\end{equation}
where $s$ is a weight-scaling factor
\begin{equation}
s = \frac{\norm{\mathbf{w}}_2}{mean(\mathbf{w})}.    
\end{equation}
Observe that  the value of $s$
always exists (because its denominator is non-zero),
and never equals zero (because its nominator is non-zero).
It is because $\norm{\mathbf{w}}_2 > 0$ and $mean(\mathbf{w}) > 0$ are guaranteed by $\mathbf{w} \neq \mathbf{0}$ (implied by the assumptions concerning $\mathbf{w}$).
In particular, for $\mathbf{w} = \mathbf{1}$, $s$ becomes:
\begin{equation}
  s = \frac{\norm{\mathbf{w}}_2}{mean(\mathbf{w})} = \frac{\norm{\mathbf{1}}_2}{mean(\mathbf{1})} = \frac{\sqrt{n}}{1} =
\sqrt{n}.
\end{equation}
which is the value of the divisor in the definition of re-scaled Euclidean distance $\delta_2^{01}$.
As a result $\delta^{01}_{\mathbf{w}}$ is a generalization and thus a full analog of the re-scaled Euclidean distance $\delta^{01}_2$.

Thanks to $s$, given any $\mathbf{w}$ and any $\mathbf{v_{1}}, \mathbf{v_{2}} \in \mathit{VS}$:
\begin{equation}
  \delta^{01}_{\mathbf{w}}(\mathbf{v_{1}},\mathbf{v_{2}}) \in [0,\frac{\norm{\mathbf{w}}_2}{s}] = [0,\frac{\norm{\mathbf{w}}_2}{\frac{\norm{\mathbf{w}}_2}{mean(\mathbf{w})}}] = [0,mean(\mathbf{w})],  
\end{equation}
 as opposed to $\delta_2(\mathbf{v_{1}},\mathbf{v_{2}}) \in [0,\norm{\mathbf{w}}_2]$. 
 Therefore, the maximal re-scaled weighted Euclidean distance $\delta^{01}_{\mathbf{w}}$  in $\mathit{VS}$ is independent of at least some characteristics of $\mathbf{w}$.
 
 Notice that the assumption `max is 1' ensures $0 < mean(\mathbf{w}) \leq 1$, so $[0,mean(\mathbf{w})]$ is a proper interval, additionally satisfying $[0,mean(\mathbf{w})] \subseteq [0,1]$.
It is also clear that for $\mathbf{w} = \mathbf{1}$, in which case $s = \sqrt{n}$, $\delta^{01}_{\mathbf{w}}$ becomes $\delta^{01}_2$: $\delta^{01}_{\mathbf{w}}(\mathbf{v_{1}},\mathbf{v_{2}}) = \delta^{01}_\mathbf{1}(\mathbf{v_{1}},\mathbf{v_{2}}) = \frac{\delta_2(\mathbf{v_{1}},\mathbf{v_{2}})}{\sqrt{n}} = \delta^{01}_2(\mathbf{v_{1}},\mathbf{v_{2}})$.

\subsection{The IA-WMSD Property in $\mathit{VS}$}
\label{sec:IA-WMSD_property}
Given two vectors $\mathbf{a}$ and $\mathbf{b} \neq \mathbf{0}$, let us define~\citep{Meyer_00}:
\begin{itemize}
\item vector $\mathbf{a} \!\searrow\! \mathbf{b} = \frac{\mathbf{a} \cdot \mathbf{b}^T}{\norm{\mathbf{b}}_2^2}  \mathbf{b}$, the \textit{orthogonal vector projection} of $\mathbf{a}$ onto $\mathbf{b}$,
\item vector $\mathbf{a} \!\nearrow\! \mathbf{b} = \mathbf{a} - \mathbf{a} \!\searrow\! \mathbf{b}$, the \textit{orthogonal vector rejection} of $\mathbf{a}$ from $\mathbf{b}$.
\end{itemize}
Notice that $\norm{\mathbf{b}}_2 \neq 0$ is guaranteed by $\mathbf{b} \neq \mathbf{0}$, so the projection vector always exists, and this means that also the rejection vector always exists.
By definition, vectors $\mathbf{a} \!\searrow\! \mathbf{b}$ and $\mathbf{a} \!\nearrow\! \mathbf{b}$ are orthogonal, i.e. $(\mathbf{a} \!\searrow\! \mathbf{b})(\mathbf{a} \!\nearrow\! \mathbf{b})^T = 0$.

Now, recall that $s = \frac{\norm{\mathbf{w}}_2}{mean(\mathbf{w})}$ for any weight vector $\mathbf{w}$. Given any $\mathbf{v} = \mathbf{u} \circ \mathbf{w} \in \mathit{VS}$ we define:
\begin{align}
mean^{01}_{\mathbf{w}}(\mathbf{v}) &= \frac{\norm{\mathbf{v} \!\searrow\! \mathbf{w}}_2}{s}, \text{which we denote as the \textit{weight-scaled mean} (WM) of $\mathbf{v}$,}\\
std^{01}_{\mathbf{w}}(\mathbf{v}) &= \frac{\norm{\mathbf{v} \!\nearrow\! \mathbf{w}}_2}{s}, \text{which we denote as the \textit{weight-scaled standard deviation} (WSD) of $\mathbf{v}$.}
\end{align}
Notice that both $mean^{01}_{\mathbf{w}}(\mathbf{v})$ and $std^{01}_{\mathbf{w}}(\mathbf{v})$ always exist, which is guaranteed by the existence of $\mathbf{v} \!\searrow\! \mathbf{w}$ and $\mathbf{v} \!\nearrow\! \mathbf{w}$ and by the fact that $s \neq 0$. Moreover, $mean^{01}_{\mathbf{w}}(\mathbf{0}) = 0$ and $mean^{01}_{\mathbf{w}}(\mathbf{w}) = mean(\mathbf{w})$, whereas
$std^{01}_{\mathbf{w}}(\mathbf{0}) = 0$ and $std^{01}_{\mathbf{w}}(\mathbf{w}) = 0$.

\begin{figure*}[htb]
\centering
\includegraphics[width=0.92\textwidth]  {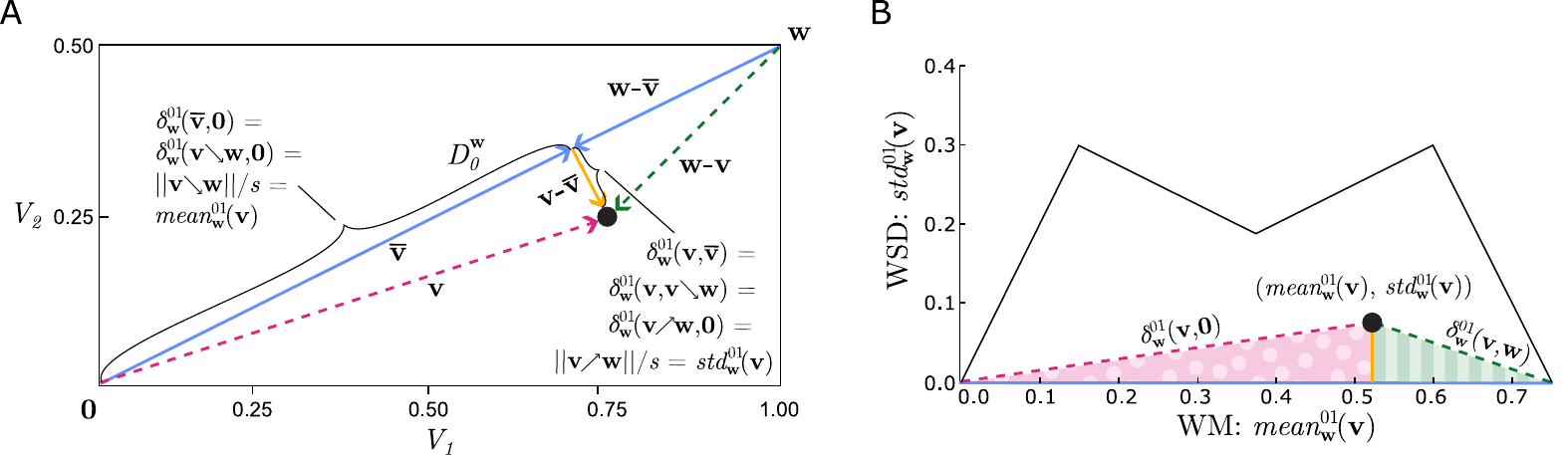}
\caption{An illustration of the IA-WMSD property in (A) $\mathit{VS}$ and (B) WMSD-space, for $\mathbf{w} = [1.0, 0.5]$ and a point $\mathbf{v} = [0.75, 0.25]$. The illustration shows how the re-scaled lengths $\delta^{01}_{\mathbf{w}}$ of vectors $\mathbf{\overline{v}}$ and $\mathbf{v} - \mathbf{\overline{v}}$ are equal to the weight-scaled mean (WM) and standard deviation (WSD), which define the WMSD-space. The diagonal $D_{\mathbf{0}}^{\mathbf{w}}$ is the blue line segment between vertices $\mathbf{0}$ and $\mathbf{w}$.}
\label{fig:IA-WMSD}
\end{figure*}

Now, given $\mathbf{v} \in \mathit{VS}$, let us observe how the diagonal $D^{\mathbf{w}}_{\mathbf{0}}$ relates $mean^{01}_{\mathbf{w}}(\mathbf{v})$ and $std^{01}_{\mathbf{w}}(\mathbf{v})$: $mean^{01}_{\mathbf{w}}(\mathbf{v})$ specifies how far away $\mathbf{v}$ is from $\mathbf{0}$ when measured \emph{along} $D^{\mathbf{w}}_{\mathbf{0}}$, while $std^{01}_{\mathbf{w}}(\mathbf{v})$ specifies how far away $\mathbf{v}$ is from $D^{\mathbf{w}}_{\mathbf{0}}$ when measured along a direction that is \emph{perpendicular} to it (Figure~\ref{fig:IA-WMSD}A). 
Thus, the diagonal can serve as a `tool' that connects ideal/anti-ideal distance calculation with the weight-scaled means and standard deviations of the alternatives.
More formally, let $\mathbf{\overline{v}}$ be defined as $\mathbf{v} \!\searrow\! \mathbf{w}$.
As visually explained in Figure~\ref{fig:IA-WMSD}A, in this case:
\begin{align}  
\delta^{01}_{\mathbf{w}}(\mathbf{\overline{v}},\mathbf{0}) &= mean^{01}_{\mathbf{w}}(\mathbf{v}),\\
\delta^{01}_{\mathbf{w}}(\mathbf{\overline{v}},\mathbf{w}) &= mean(\mathbf{w}) - mean^{01}_{\mathbf{w}}(\mathbf{v}),\\
\delta^{01}_{\mathbf{w}}(\mathbf{\overline{v}},\mathbf{v}) &= std^{01}_{\mathbf{w}}(\mathbf{v}).
\end{align}

Additionally, because $\mathbf{v} = \mathbf{w} \circ \mathbf{u}$ and for $\mathbf{w} = \mathbf{1}$ we get $s = \sqrt{n}$, it may be shown that: 
\begin{align} 
mean^{01}_{\mathbf{1}}(\mathbf{v}) &= 
\frac{\norm{\mathbf{v} \!\searrow\! \mathbf{1}}_2}{\sqrt{n}} = 
\frac{\norm{(\mathbf{1} \circ \mathbf{u}) \!\searrow\! \mathbf{1}}_2}{\sqrt{n}} = 
\frac{\norm{\mathbf{u} \!\searrow\! \mathbf{1}}_2}{\sqrt{n}} = 
\frac{\frac{\mathbf{u} \cdot \mathbf{1}^T}{\norm{\mathbf{1}}_2^2} \cdot \; \norm{\mathbf{1}}_2}{\sqrt{n}} = \notag\\& = 
\frac{\frac{\mathbf{u} \cdot \mathbf{1}^T}{\norm{\mathbf{1}}_2}}{\sqrt{n}} = 
\frac{\frac{\mathbf{u} \cdot \mathbf{1}^T}{\sqrt{n}}}{\sqrt{n}} = \frac{\mathbf{u} \cdot \mathbf{1}^T}{n} = \frac{sum(\mathbf{u})}{n} = mean(\mathbf{u}),\\
std^{01}_{\mathbf{1}}(\mathbf{v}) &= 
\frac{\norm{\mathbf{v} \!\nearrow\! \mathbf{1}}_2}{\sqrt{n}} = 
\frac{\norm{(\mathbf{1} \circ \mathbf{u}) \!\nearrow\! \mathbf{1}}_2}{\sqrt{n}} = 
\frac{\norm{\mathbf{u} \!\nearrow\! \mathbf{1}}_2}{\sqrt{n}} = \notag\\&
= \frac{\norm{\mathbf{u} - \mathbf{u} \!\searrow\! \mathbf{1}}_2}{\sqrt{n}} =
\frac{\norm{\mathbf{u} - \frac{\mathbf{u} \cdot \mathbf{1}^T}{\norm{\mathbf{1}}_2^2} \cdot \mathbf{1}}_2}{\sqrt{n}} =
\frac{\norm{\mathbf{u} - \frac{sum(\mathbf{u})}{\sqrt{n}^2} \cdot \mathbf{1}}_2}{\sqrt{n}} = \notag\\&=
\frac{\norm{\mathbf{u} - \frac{sum(\mathbf{u})}{n} \cdot \mathbf{1}}_2}{\sqrt{n}} =
\frac{\norm{\mathbf{u} - mean(\mathbf{u}) \cdot \mathbf{1}}_2}{\sqrt{n}} =
\frac{\sqrt{(\mathbf{u} - mean(\mathbf{u}) \cdot \mathbf{1})(\mathbf{u} - mean(\mathbf{u}) \cdot \mathbf{1})^T}}{\sqrt{n}} = \notag\\&=
\sqrt{\frac{(\mathbf{u} - mean(\mathbf{u}) \cdot \mathbf{1})(\mathbf{u} - mean(\mathbf{u}) \cdot \mathbf{1})^T}{n}} =
\sqrt{var(\mathbf{u})} =
std(\mathbf{u}),
\end{align}
which means that $mean^{01}_{\mathbf{w}}(\mathbf{v})$ and $std^{01}_{\mathbf{w}}(\mathbf{v})$ constitute natural generalizations of $mean(\mathbf{u})$ and $std(\mathbf{u})$.

As can be noticed in Figure~\ref{fig:IA-WMSD}A, $mean^{01}_{\mathbf{w}}(\mathbf{v})$ satisfies:
\begin{equation}
 mean^{01}_{\mathbf{w}}(\mathbf{v}) = \delta^{01}_{\mathbf{w}}(\mathbf{\overline{v}},\mathbf{0}) \equiv \delta^{01}_{\mathbf{w}}(\mathbf{v} \!\searrow\! \mathbf{w},\mathbf{0}).   
\end{equation}
 Simultaneously, $std^{01}_{\mathbf{w}}(\mathbf{v})$ satisfies:
 \begin{equation}
   std^{01}_{\mathbf{w}}(\mathbf{v}) = \delta^{01}_{\mathbf{w}}(\mathbf{\overline{v}},\mathbf{v}) \equiv \delta^{01}_{\mathbf{w}}(\mathbf{v} \!\searrow\! \mathbf{w},\mathbf{v}) = \delta^{01}_{\mathbf{w}}(\mathbf{v} \!\nearrow\! \mathbf{w},\mathbf{0}). 
 \end{equation}
 All of the abovementioned considerations allow us to formulate the IA-WMSD property.

\begin{theorem}[IA-WMSD Property]
For every $\mathbf{w}$ defining $\mathbf{v} \in \mathit{VS}$:
\hfill
\begin{align}
\delta^{01}_{\mathbf{w}}(\mathbf{v},\mathbf{0}) &= 
\sqrt{mean^{01}_\mathbf{w}(\mathbf{v})^2+std^{01}_\mathbf{w}(\mathbf{v})^2}, \\
\delta^{01}_{\mathbf{w}}(\mathbf{v},\mathbf{w}) &= 
\sqrt{\big(mean(\mathbf{w})-mean^{01}_\mathbf{w}(\mathbf{v})\big)^2+std^{01}_\mathbf{w}(\mathbf{v})^2}.
\end{align}
\end{theorem}
\noindent
Notice that $mean(\mathbf{w})$ in the above may also be expressed as: 
\begin{equation}
    mean(\mathbf{w}) = 
\frac{\norm{\mathbf{w}}_2}{\norm{\mathbf{w}}_2} \cdot mean(\mathbf{w}) = \frac{\norm{\mathbf{w}}_2}{\frac{\norm{\mathbf{w}}_2}{mean(\mathbf{w})}} = \frac{\norm{\mathbf{w}}_2}{s},
\end{equation}
which emphasizes the divisor $s$, common to the core definitions of this paper. It should be also additionally stressed that for $\mathbf{w} = \mathbf{1}$ the IA-WMSD property becomes the IA-MSD property.
In the next section, we will discuss how the IA-WMSD property can be used to create $n$-independent visualizations of weight-based TOPSIS aggregations.

\subsection{The WMSD-space}
\label{sec:WMSD-space}
Analogously to the case of unweighted criteria and the resulting MSD-space~\citep{MSD-space}, the relation between the re-scaled weighted distances of an alternative and the predefined reference points allows us to propose a new space called \textit{WMSD-space} that uses $mean^{01}_{\mathbf{w}}(\mathbf{v})$ (WM) and $std^{01}_{\mathbf{w}}(\mathbf{v})$ (WSD) as its components.

\begin{definition}[WMSD-space]
\hfill
\begin{align}
\text{WMSD-space} =\{[mean^{01}_{\mathbf{w}}(\mathbf{v}),std^{01}_{\mathbf{w}}(\mathbf{v})] \; | \; \mathbf{v} \in VS\}
\end{align}
\end{definition}

The WMSD-space can be represented in 2D space wherein the weight-scaled mean (WM) of the alternatives is presented on the x-axis and the weight-scaled standard deviation (WSD) of the alternatives on the y-axis. As depicted in Figure~\ref{fig:IA-WMSD}, WMSD-space may be treated as an image of $\mathit{VS}$ under a two-dimensional transformation by functions
$mean^{01}_{\mathbf{w}}(\mathbf{v})$ and $std^{01}_{\mathbf{w}}(\mathbf{v})$.
It is worth underlining that the IA-WMSD property holds in WMSD-space, where it follows the Pythagorean theorem for two right triangles (pink dotted and green striped triangles in Figure~\ref{fig:IA-WMSD}B). This is a result of WMSD-space being a `rotational' projection of $\mathit{VS}$ into two dimensions that retain the IA-WMSD property.

\begin{figure*}[h!]
\centering
\includegraphics[width=0.98\textwidth]  {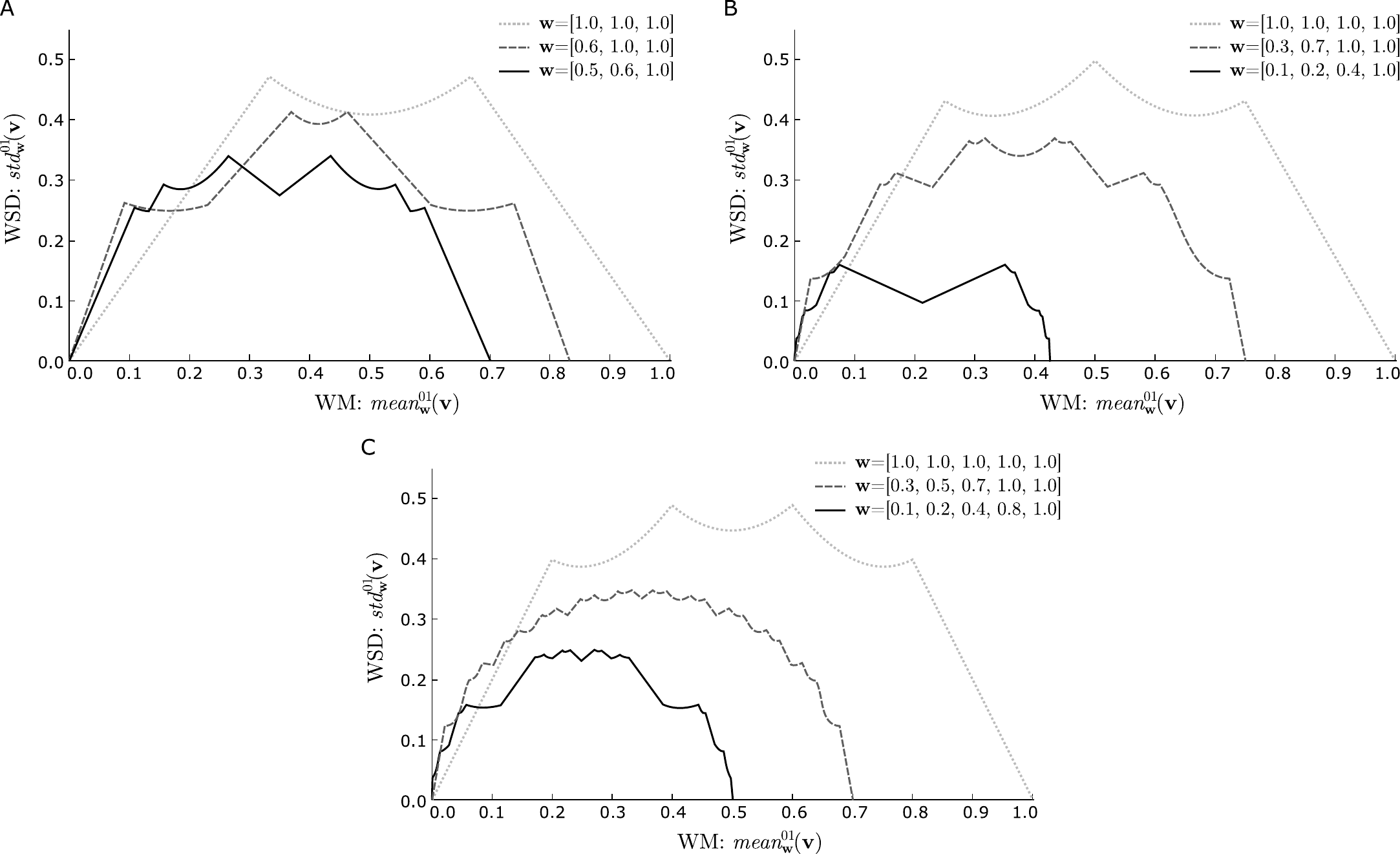}
\caption{Visualizations of WMSD-space for the number of criteria (A) $n=3$, (B) $n=4$, (C) $n=5$, each for three different sets of weights depicted by different line types. Notice that the dotted light gray line on each subplot corresponds to uniform weights and, therefore, the special case of WMSD-space, which is MSD-space. It is also worth noting how the arithmetic mean of the weight ($mean(\mathbf{w})$) corresponds to the maximal x-axis coordinate of WMSD-space.}
\label{fig:WMSD}
\end{figure*}

Since WMSD-space is actually based on the weighted utility space, which is bounded, the extreme values of WM and WSD are also bounded. In other words, for a given set of criteria weights, there is only a limited range of attainable WM and WSD values\footnote{Additional remarks on the boundary are presented in \ref{app:perimeters_of_WMSD-space}.}. As a result, one can depict the boundary of WMSD-space, which depends on $\mathbf{w}$, and thus also on the number of criteria $n_p$. Figure~\ref{fig:WMSD} presents the shape of WMSD-space for different numbers of criteria $n$ and weight vectors $\mathbf{w}$.
Owing to the symmetry of $\mathit{VS}$ and its `rotational' projections, the boundary of the WMSD-space does not depend on the order of the weights of $\mathbf{w}$, which means that the shape of the WMSD-space remains the same for every permutation of a given set of weights.

Looking at Figure~\ref{fig:WMSD} it can be noticed that setting $\mathbf{w} = \mathbf{1}$ makes WMSD-space equivalent to MSD-space. Indeed, the weight-scaling factor $s = \frac{\norm{\mathbf{w}}_2}{mean(\mathbf{w})}$ has been chosen to replicate the relation between $\mathit{VS}$ and $\mathit{US}$ in the relation between WMSD-space and MSD-space. Let us recall that MSD-space is based on a re-scaled distance measure $\delta^{01}_2$, which simply divides the Euclidean distance $\delta_2$ by $\sqrt{n}$. By doing so, $\delta^{01}_2$ is independent of the number of criteria $n$, making the maximum distance in MSD-space always 1. Notice that $\sqrt{n}$ is a special case of $\norm{\mathbf{w}}_2$ for $\mathbf{w} = \mathbf{1}$, which emphasizes the fact that $\delta^{01}_{\mathbf{w}}$ is a natural generalization of $\delta^{01}_2$ for the case of $\mathbf{w} \neq \mathbf{1}$. Although making $s = \norm{\mathbf{w}}_2$ would suffice to ensure the IA-WMSD property, additionally scaling by $mean(\mathbf{w})$ makes the sizes between WMSD-space and MSD-space follow the relation $mean(\mathbf{w}) < mean(\mathbf{1}) = 1$ for $\mathbf{w} \neq \mathbf{1}$. As a result, in WMSD-space the maximal value of WM is $mean(\mathbf{w})$ instead of 1.

It is also worth noticing that the number of non-zero criteria and the particular values of their weights (i.e. the size and the values of $\mathbf{w}$) affect the number of vertices of the WMSD-space boundary (Figure~\ref{fig:WMSD}). Finally, as was the case for MSD-space, WMSD-space can always be depicted in two dimensions. We will use this property to visualize alternatives and values of TOPSIS aggregation functions in WMSD-space.

\subsection{TOPSIS Aggregations in WMSD-space}
\label{sec:visualization}
The application of the weights, while changing $\mathit{US}$ into $\mathit{VS}$, does necessarily influence the image of the ideal point, as the image moves from $\mathbf{1}$ in $\mathit{US}$ to $\mathbf{w}$ in $\mathit{VS}$. The same does not concern the image of the anti-ideal point, as it remains the same, being equal to $\mathbf{0}$ in $\mathit{US}$ and to $\mathbf{w} \circ \mathbf{0} = \mathbf{0}$ in $\mathit{VS}$. 
As a result, TOPSIS utilizes $\mathbf{w}$ instead of $\mathbf{1}$ when computing the `distance to the ideal'. This means that new versions of the aggregation functions denoted
by $\mathsf{I}_{\mathbf{w}}$, $\mathsf{A}_{\mathbf{w}}$ and $\mathsf{R}_{\mathbf{w}}$, must be introduced.
When expressed in terms of $\delta^{01}_{\mathbf{w}}(\mathbf{v},\mathbf{w})$ (`distance to the ideal in $\mathit{VS}$') and $\delta^{01}_{\mathbf{w}}(\mathbf{v},\mathbf{0})$ (`distance to the anti-ideal in $\mathit{VS}$'), they are defined as follows:
\begin{align} 
\mathsf{I}_{\mathbf{w}}(\mathbf{v}) &= 1 - \frac{\delta^{01}_{\mathbf{w}}(\mathbf{v},\mathbf{w})}{mean(\mathbf{w})}, \\
\mathsf{A}_{\mathbf{w}}(\mathbf{v}) &= \frac{\delta^{01}_{\mathbf{w}}(\mathbf{v},\mathbf{0})}{mean(\mathbf{w})}, \\
\mathsf{R}_{\mathbf{w}}(\mathbf{v}) &= \frac{\delta^{01}_{\mathbf{w}}(\mathbf{v},\mathbf{0})}{\delta^{01}_{\mathbf{w}}(\mathbf{v},\mathbf{w})+\delta^{01}_{\mathbf{w}}(\mathbf{v},\mathbf{0})}.
\end{align}
where $\mathbf{u} \in \mathit{US}$ and $\mathbf{v} = \mathbf{w} \circ \mathbf{u} \in \mathit{VS}$ are the images of an alternative from $\mathit{US}$ and $\mathit{VS}$, respectively.
All three aggregations generate values belonging to $[0, 1]$.

As was the case with $\mathsf{I}(\mathbf{u})$ (see Section~\ref{sec:preliminaries}), aggregation $\mathsf{I}_\mathbf{w}(\mathbf{v})$ features a reversal ($1 - \frac{\delta^{01}_{\mathbf{w}}(\mathbf{v},\mathbf{w})}{mean(\mathbf{w})}$ instead of $\frac{\delta^{01}_{\mathbf{w}}(\mathbf{v},\mathbf{w})}{mean(\mathbf{w})}$). This modification was introduced to ensure that all aggregations are interpreted as functions that need to be maximized. The distances from the ideal and anti-ideal points in $\mathsf{I}_{\mathbf{w}}(\mathbf{v})$ and $\mathsf{A}_{\mathbf{w}}(\mathbf{v})$ have been divided by $mean(\mathbf{w})$ to keep the values of these aggregations in $[0, 1]$.

Obviously, using the IA-WMSD property it is possible to express all the aggregations in terms of $mean^{01}_\mathbf{w}(\mathbf{v})$ and $std^{01}_\mathbf{w}(\mathbf{v})$ instead of $\delta^{01}_\mathbf{w}(\mathbf{v},\mathbf{w})$ and $\delta^{01}_\mathbf{w}(\mathbf{v},\mathbf{0})$:

\begin{align} 
\mathsf{I}_{\mathbf{w}}(\mathbf{v}) &= 1 - \frac{\sqrt{\big(mean(\mathbf{w})-mean^{01}_\mathbf{w}(\mathbf{v})\big)^2+std^{01}_\mathbf{w}(\mathbf{v})^2}}{mean(\mathbf{w})}, \\
\mathsf{A}_{\mathbf{w}}(\mathbf{v}) &= \frac{\sqrt{mean^{01}_\mathbf{w}(\mathbf{v})^2+std^{01}_\mathbf{w}(\mathbf{v})^2}}{mean(\mathbf{w})}, \\
\mathsf{R}_{\mathbf{w}}(\mathbf{v}) &= \frac{\sqrt{mean^{01}_\mathbf{w}(\mathbf{v})^2+std^{01}_\mathbf{w}(\mathbf{v})^2}}{\sqrt{mean^{01}_\mathbf{w}(\mathbf{v})^2+std^{01}_\mathbf{w}(\mathbf{v})^2}+\sqrt{\big(mean(\mathbf{w})-mean^{01}_\mathbf{w}(\mathbf{v})\big)^2+std^{01}_\mathbf{w}(\mathbf{v})^2}}.\quad \quad \quad
\end{align}

Notice that each aggregation is a function of a single vector argument $\mathbf{v} \in VS$, and thus, in practice, a function of the $n$ real-valued elements of this vector. However, the same IA-WMSD property makes it possible to render these aggregations in a plane as a function of just two real-valued arguments: $mean^{01}_\mathbf{w}(\mathbf{v})$ and $std^{01}_\mathbf{w}(\mathbf{v})$. This demonstrates the introduced underlying dimensionality reduction. 
For step-by-step examples of how the equations presented in the above sections can be used to transform an alternative from the criteria space to WMSD-space see~\ref{app:detailed_example}.

Now, when visualizing WMSD-space it is helpful to show alternatives against the values of TOPSIS aggregations ($\mathsf{I}_\mathbf{w}(\mathbf{v})$, $\mathsf{A}_\mathbf{w}(\mathbf{v})$ and $\mathsf{R}_\mathbf{w}(\mathbf{v})$), which can be color-coded as was done for MSD-space~\citep{MSD-space}. 

\begin{figure*}[h!]
\centering
\includegraphics[width=0.98\textwidth]  {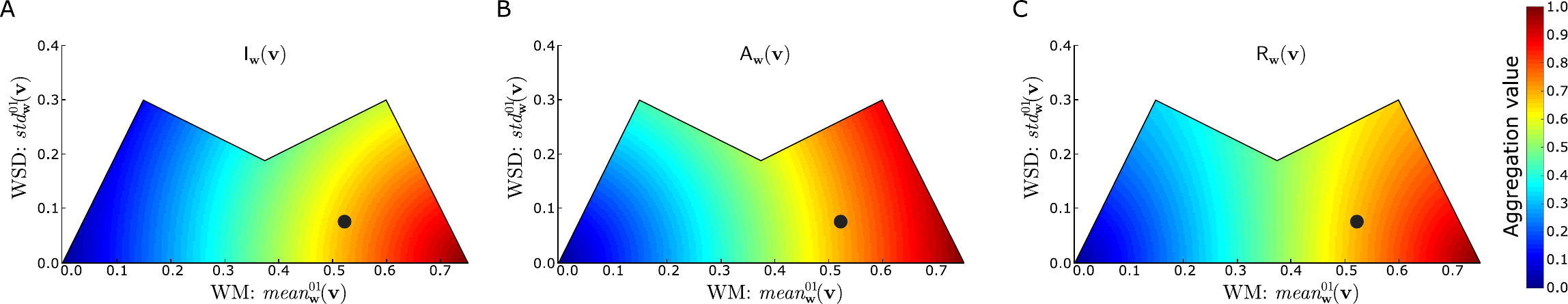}
\caption{An illustrative point $\mathbf{v} = [0.75, 0.25]$ depicted in WMSD-space defined by $\mathbf{w} = [1.0, 0.5]$  for aggregations (A) $\mathsf{I}_\mathbf{w}(\mathbf{v})$, (B) $\mathsf{A}_\mathbf{w}(\mathbf{v})$ and (C) $\mathsf{R}_\mathbf{w}(\mathbf{v})$. Color encodes the aggregation value, with blue representing the least preferred and red the most preferred values.}
\label{fig:WMSD-colors}
\end{figure*}


As seen in Figure~\ref{fig:WMSD-colors}, \emph{coloring WMSD-space in a way that represents the values of the aggregation functions reveals the interplay of $mean^{01}_\mathbf{w}(\mathbf{v})$ and $std^{01}_\mathbf{w}(\mathbf{v})$.} The WM-WSD interplay for different aggregation functions in WMSD-space resembles that described for MSD-space. This is to be expected as MSD-space constitutes a special case of WMSD-space where $\mathbf{w} = \mathbf{1}$. Table~\ref{tab:preference-related-interplay}, shows which aggregation functions act like type cost or type gain criteria depending on $mean^{01}_\mathbf{w}(\mathbf{v})$ and $std^{01}_\mathbf{w}(\mathbf{v})$, and compares WMSD-space with MSD-space.

\begin{table}[htb]
  \centering
  \caption{The relation between the mean and standard deviation as the property of the analyzed aggregation functions. On the left, properties of aggregation functions in MSD-space, as presented in~\cite{MSD-space}. On the right, the properties of aggregation functions in WMSD-space. Notably, the configuration of gain/cost table entries is the same on the left and on the right.}
    \begin{tabular}{cccccc}
    \toprule
    \multirow{2}{*}{aggregation} & \multicolumn{2}{c}{MSD-space}  & & \multicolumn{2}{c}{WMSD-space}  \\ 
    \cline{2-3} \cline{5-6} \\[-0.9em]
     & $mean(\mathbf{u})$ & $std(\mathbf{u})$ & & $mean^{01}_\mathbf{w}(\mathbf{v})$ & $std^{01}_\mathbf{w}(\mathbf{v})$ \\
    \midrule
    $\mathsf{I}$ & gain & cost & & gain & cost \\
    &&\\
    $\mathsf{A}$ & gain & gain & & gain & gain \\
    &&\\
    \multirow{3}{*}{$\mathsf{R}$} & \multirow{3}{*}{gain} & \multicolumn{1}{l}{$mean(\mathbf{u}) < 0.5$: gain} & & \multirow{3}{*}{gain}  & \multicolumn{1}{l}{$mean^{01}_\mathbf{w}(\mathbf{v}) < \frac{mean(\mathbf{w})}{2}$: gain}\\
      & & \multicolumn{1}{l}{$mean(\mathbf{u}) = 0.5$: neutrality} & & & \multicolumn{1}{l}{$mean^{01}_\mathbf{w}(\mathbf{v}) = \frac{mean(\mathbf{w})}{2}$: neutrality}\\
      & & \multicolumn{1}{l}{$mean(\mathbf{u}) > 0.5$: cost} & & & \multicolumn{1}{l}{$mean^{01}_\mathbf{w}(\mathbf{v}) > \frac{mean(\mathbf{w})}{2}$: cost}\\
    \bottomrule
    \end{tabular}%
  \label{tab:preference-related-interplay}%
\end{table}%

As can be seen in Table~\ref{tab:preference-related-interplay}, MSD-space and WMSD-space share the same configuration of gain/cost characteristics of the means and standard deviations across the spaces. It is also worth noticing that the mean weight $mean(\mathbf{w})$ plays a role in the properties of aggregation  $\mathsf{R}_\mathbf{w}(\mathbf{v})$. Finally, it can be observed that WMSD-space is a generalization of MSD-space to weighted criteria ($\mathbf{v} = \mathbf{w} \circ \mathbf{u}$). For the special case of $\mathbf{w} = \mathbf{1}$, WMSD-space is equivalent to MSD-space (in particular, $\frac{mean(\mathbf{1})}{2} = 0.5$). In the following section, we will discuss how these properties apply to practical ranking problems.

\section{Case Studies}
\label{sec:case}
In this section, we present two case studies conducted on a dataset of students described in terms of school grades and on a dataset of countries described in terms of factors constituting the Index of Economic Freedom. The goal of the case studies is to visualize the alternatives within the MSD-space and WMSD-space, present the impact of introducing weights, and discuss how the two spaces depict the relations between each alternative’s properties and their aggregation values. The datasets and code for reproducing the results presented in this section are available at GitHub: \url{https://github.com/dabrze/topsis-msd-improvement-actions}.

In the next subsections, the following notation shall be used to present alternative rankings: $\mathbf{X}_{i} \succ_{agg} \mathbf{X}_{j}$: alternative $\mathbf{X}_{i}$ is preferred over $\mathbf{X}_{j}$ under every aggregation from $agg$; $\mathbf{X}_{i} \sim_{agg} \mathbf{X}_{j}$: $\mathbf{X}_{i}$ and $\mathbf{X}_{j}$ are indifferent under $agg$; $\mathbf{X}_{i} \prec_{agg} \mathbf{X}_{j}$: $\mathbf{X}_{j}$ is preferred over $\mathbf{X}_{i}$ under $agg$. Moreover, for an alternative $\mathbf{X}_i$ we will use $mean(\mathbf{X}_i)$ as a shorthand for $mean(\mathit{US}(\mathbf{X}_i))$ and $mean^{01}_\mathbf{w}(\mathbf{X}_i)$ as a shorthand for $mean^{01}_\mathbf{w}(\mathit{VS}(\mathbf{X}_i))$. Analogously, $std(\mathbf{X}_i)$ and $std^{01}_\mathbf{w}(\mathbf{X}_i)$ will also denote re-scaled standard deviations calculated for images of the alternative in $\mathit{US}$ and $\mathit{VS}$ respectively. 

\subsection{Student Grades}
\label{sec:grades}
\begin{figure*}[htbp]
\centering
\includegraphics[width=0.98\textwidth]{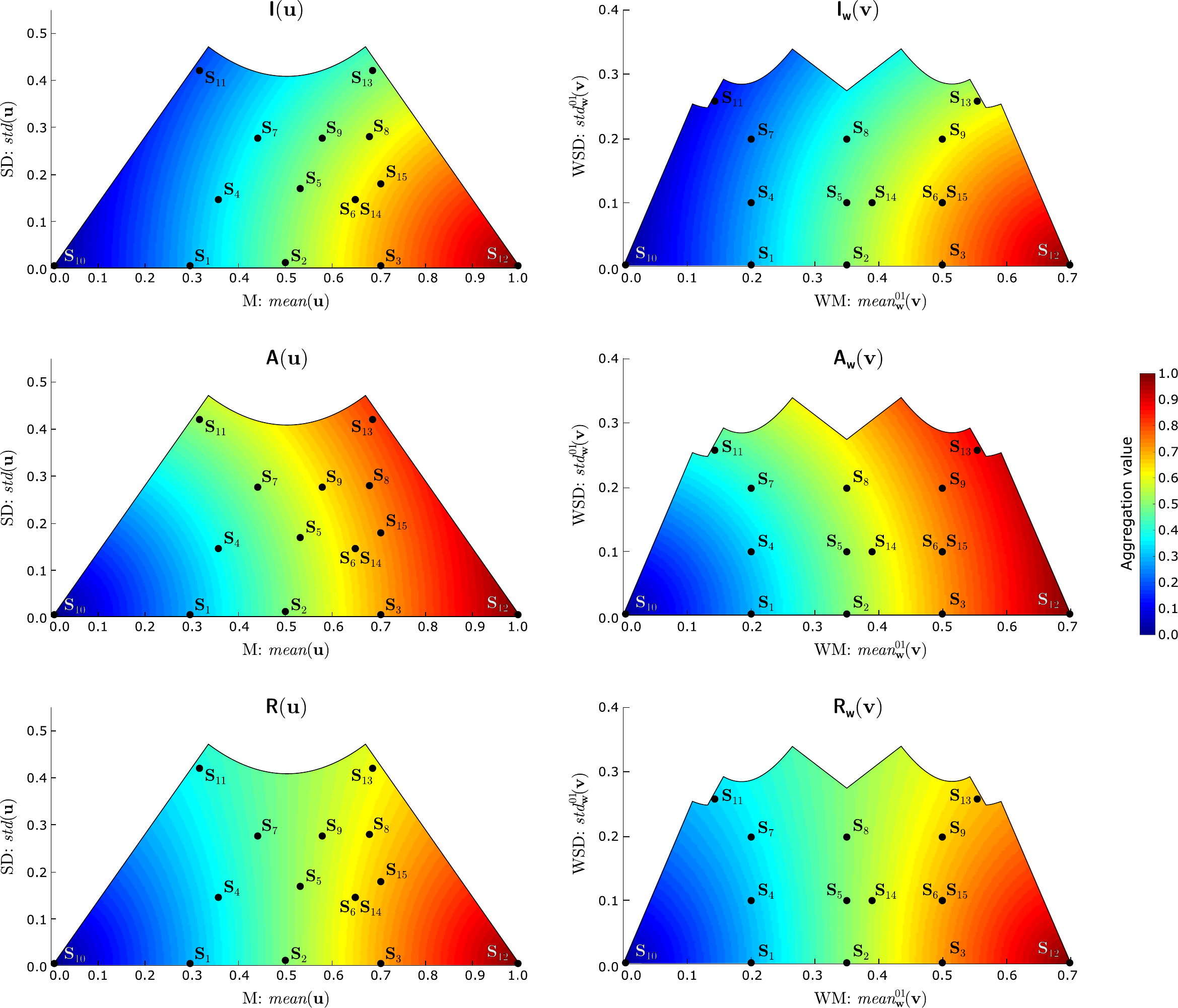}
\caption{Students depicted in MSD-space (left) and WMSD-space defined for $\mathbf{w}=[0.5, 0.6, 1.0]$ (right) for three different aggregation functions: $\mathsf{I}(\mathbf{u})$ (top), $\mathsf{A}(\mathbf{u})$ (middle) and $\mathsf{R}(\mathbf{u})$ (bottom). Color encodes the aggregation value, with blue representing the least preferred and red the most preferred values. Note that the WMSD sub-figures on the right are on a different scale than those on the left; we have scaled up the WMSD sub-figures to increase their readability.}
\label{fig:sandcastle3-weightcastle3-05-06-1-I-A-T}
\end{figure*}

The first dataset contains 15 alternatives, i.e., students described by three criteria which are the average grades obtained by these students in Maths, Biology, and Art. The domains of the criteria are $[0,100]$ for Maths, $[1,6]$ for Biology, and $[1,6]$ for Art. The alternatives are presented in Table~\ref{tab:students}.
The rankings of the alternatives are considered in two scenarios: when all criteria are of equal importance (i.e., $\mathbf{w} = [1.0, 1.0, 1.0]$) and when $\mathbf{w} = [0.5, 0.6, 1.0]$. The description of alternatives in terms of $\mathit{US}$ (equal weights), $\mathit{VS}$ (unequal weights), MSD-space, WMSD-space and the three aggregations are presented in Table~\ref{tab:exemplary-alternatives}.
The alternatives have been chosen to cover different areas of the MSD-space and WMSD-space (see~Figure~\ref{fig:sandcastle3-weightcastle3-05-06-1-I-A-T}) and to represent some of their characteristic points, e.g. the worst possible alternative ($\mathbf{S}_{10}$) or best possible alternative ($\mathbf{S}_{12}$). 

\begin{table}[htbp]
    \centering
  \caption{Descriptions of alternatives for the first case study; students described by their average grades from Maths $[0-100]$, Biology $[1-6]$, and Art $[1-6]$.}
  \label{tab:students}
    \begin{tabular}{lrrr}
    \toprule
    { } & Math & Bio & Art \\
    \midrule
        $\mathbf{S}_{1}$ & 29.11 & 2.46 & 2.46 \\ 
        $\mathbf{S}_{2}$ & 49.37 & 3.53 & 3.47 \\ 
        $\mathbf{S}_{3}$ & 70.89 & 4.54 & 4.54 \\ 
        $\mathbf{S}_{4}$ & 40.51 & 3.53 & 1.89 \\ 
        $\mathbf{S}_{5}$ & 35.44 & 4.80 & 3.22 \\ 
        $\mathbf{S}_{6}$ & 59.49 & 3.47 & 5.11 \\ 
        $\mathbf{S}_{7}$ & 44.30 & 4.80 & 1.38 \\ 
        $\mathbf{S}_{8}$ & 93.67 & 5.05 & 2.39 \\ 
        $\mathbf{S}_{9}$ & 55.70 & 2.20 & 5.62 \\ 
        $\mathbf{S}_{10}$ & 0.00 & 1.00 & 1.00 \\ 
        $\mathbf{S}_{11}$ & 0.00 & 5.56 & 1.13 \\ 
        $\mathbf{S}_{12}$ & 100.00 & 6.00 & 6.00 \\ 
        $\mathbf{S}_{13}$ & 100.00 & 1.44 & 5.87 \\ 
        $\mathbf{S}_{14}$ & 70.71 & 4.84 & 3.22 \\ 
        $\mathbf{S}_{15}$ & 89.90 & 3.32 & 4.79 \\
    \bottomrule
    \end{tabular}
\end{table}

\begin{table*}[htbp]
\footnotesize
  \centering
  \caption{Descriptions of students in terms of $\mathit{US}$, $\mathit{VS}$ given $\mathbf{w} = [0.5, 0.6, 1.0]$, MSD-space, WMSD-space and the three aggregations. The ranking positions of alternatives for particular aggregations are presented as superscripts $^{(\texttt{pos})}$.}
    \begin{tabular}{@{}l@{\hspace{1.25em}}r@{\hspace{0.5em}}r@{\hspace{0.5em}}r@{\qquad}r@{\hspace{0.5em}}r@{\hspace{0.5em}}r@{\qquad}r@{\hspace{0.5em}}r@{\qquad}r@{\hspace{0.5em}}r@{\qquad}l@{\hspace{0.5em}}l@{\hspace{0.5em}}ll@{\hspace{0.5em}}l@{\hspace{0.5em}}l@{}}
	\toprule
         & \multicolumn{3}{c}{$\mathit{US}\qquad$} & \multicolumn{3}{c}{$\mathit{VS}\qquad$} & \multicolumn{2}{l}{MSD} & \multicolumn{2}{l}{WMSD} & \multicolumn{6}{c}{Aggregations} \\
        { } & $U_1$ & $U_2$ & $U_3$ &
        $V_1$ & $V_2$ & $V_3$ &
		M & SD  &  WM & WSD  & 
		$\mathsf{I}(\mathbf{u})$ & $\mathsf{A}(\mathbf{u})$ & $\mathsf{R}(\mathbf{u})$  & 
		$\mathsf{I_{\mathbf{w}}}(\mathbf{v})$ & $\mathsf{A_{\mathbf{w}}}(\mathbf{v})$ & $\mathsf{R_{\mathbf{w}}}(\mathbf{v})$ \\
  \midrule
        $\mathbf{S}_{1}$ & 0.29 & 0.29 & 0.29 & 0.15 & 0.17 & 0.29 & 0.29 & 0.00 & 0.20 & 0.00 & 0.29$^{(\texttt{13})}$ & 0.29$^{(\texttt{14})}$ & 0.29$^{(\texttt{14})}$ & 0.29$^{(\texttt{11})}$ & 0.29$^{(\texttt{14})}$ & 0.29$^{(\texttt{14})}$ \\ 
        $\mathbf{S}_{2}$ & 0.49 & 0.51 & 0.49 & 0.25 & 0.30 & 0.49 & 0.50 & 0.01 & 0.35 & 0.00 & 0.50$^{(\texttt{7})}$ & 0.50$^{(\texttt{12})}$ & 0.50$^{(\texttt{10})}$ & 0.50$^{(\texttt{8})}$ & 0.50$^{(\texttt{10})}$ & 0.50$^{(\texttt{8})}$ \\ 
        $\mathbf{S}_{3}$ & 0.71 & 0.71 & 0.71 & 0.35 & 0.43 & 0.71 & 0.71 & 0.00 & 0.50 & 0.00 & 0.71$^{(\texttt{2})}$ & 0.71$^{(\texttt{5})}$ & 0.71$^{(\texttt{2})}$ & 0.71$^{(\texttt{2})}$ & 0.71$^{(\texttt{6})}$ & 0.71$^{(\texttt{2})}$ \\ 
        $\mathbf{S}_{4}$ & 0.41 & 0.51 & 0.18 & 0.20 & 0.30 & 0.18 & 0.36 & 0.14 & 0.20 & 0.10 & 0.35$^{(\texttt{12})}$ & 0.39$^{(\texttt{13})}$ & 0.37$^{(\texttt{13})}$ & 0.27$^{(\texttt{12})}$ & 0.32$^{(\texttt{13})}$ & 0.31$^{(\texttt{13})}$ \\ 
        $\mathbf{S}_{5}$ & 0.35 & 0.76 & 0.44 & 0.18 & 0.46 & 0.44 & 0.52 & 0.17 & 0.35 & 0.10 & 0.49$^{(\texttt{8})}$ & 0.55$^{(\texttt{9})}$ & 0.52$^{(\texttt{9})}$ & 0.48$^{(\texttt{9})}$ & 0.52$^{(\texttt{9})}$ & 0.50$^{(\texttt{8})}$ \\ 
        $\mathbf{S}_{6}$ & 0.59 & 0.49 & 0.82 & 0.30 & 0.30 & 0.82 & 0.64 & 0.14 & 0.50 & 0.10 & 0.61$^{(\texttt{4})}$ & 0.65$^{(\texttt{6})}$ & 0.63$^{(\texttt{4})}$ & 0.68$^{(\texttt{3})}$ & 0.73$^{(\texttt{4})}$ & 0.69$^{(\texttt{3})}$ \\ 
        $\mathbf{S}_{7}$ & 0.44 & 0.76 & 0.08 & 0.22 & 0.46 & 0.08 & 0.43 & 0.28 & 0.20 & 0.20 & 0.36$^{(\texttt{11})}$ & 0.51$^{(\texttt{11})}$ & 0.44$^{(\texttt{11})}$ & 0.23$^{(\texttt{13})}$ & 0.40$^{(\texttt{12})}$ & 0.34$^{(\texttt{11})}$ \\ 
        $\mathbf{S}_{8}$ & 0.94 & 0.81 & 0.28 & 0.47 & 0.49 & 0.28 & 0.68 & 0.29 & 0.35 & 0.20 & 0.57$^{(\texttt{6})}$ & 0.73$^{(\texttt{3})}$ & 0.63$^{(\texttt{4})}$ & 0.42$^{(\texttt{10})}$ & 0.58$^{(\texttt{7})}$ & 0.50$^{(\texttt{8})}$ \\ 
        $\mathbf{S}_{9}$ & 0.56 & 0.24 & 0.92 & 0.28 & 0.14 & 0.92 & 0.57 & 0.28 & 0.50 & 0.20 & 0.49$^{(\texttt{8})}$ & 0.64$^{(\texttt{8})}$ & 0.56$^{(\texttt{8})}$ & 0.60$^{(\texttt{5})}$ & 0.77$^{(\texttt{3})}$ & 0.66$^{(\texttt{6})}$ \\ 
        $\mathbf{S}_{10}$ & 0.00 & 0.00 & 0.00 & 0.00 & 0.00 & 0.00 & 0.00 & 0.00 & 0.00 & 0.00 & 0.00$^{(\texttt{15})}$ & 0.00$^{(\texttt{15})}$ & 0.00$^{(\texttt{15})}$ & 0.00$^{(\texttt{15})}$ & 0.00$^{(\texttt{15})}$ & 0.00$^{(\texttt{15})}$ \\ 
        $\mathbf{S}_{11}$ & 0.00 & 0.91 & 0.03 & 0.00 & 0.55 & 0.03 & 0.31 & 0.42 & 0.15 & 0.26 & 0.19$^{(\texttt{14})}$ & 0.53$^{(\texttt{10})}$ & 0.39$^{(\texttt{12})}$ & 0.14$^{(\texttt{14})}$ & 0.43$^{(\texttt{11})}$ & 0.33$^{(\texttt{12})}$ \\ 
        $\mathbf{S}_{12}$ & 1.00 & 1.00 & 1.00 & 0.50 & 0.60 & 1.00 & 1.00 & 0.00 & 0.70 & 0.00 & 1.00$^{(\texttt{1})}$ & 1.00$^{(\texttt{1})}$ & 1.00$^{(\texttt{1})}$ & 1.00$^{(\texttt{1})}$ & 1.00$^{(\texttt{1})}$ & 1.00$^{(\texttt{1})}$ \\ 
        $\mathbf{S}_{13}$ & 1.00 & 0.09 & 0.97 & 0.50 & 0.05 & 0.97 & 0.69 & 0.42 & 0.55 & 0.26 & 0.47$^{(\texttt{10})}$ & 0.81$^{(\texttt{2})}$ & 0.61$^{(\texttt{7})}$ & 0.57$^{(\texttt{6})}$ & 0.86$^{(\texttt{2})}$ & 0.67$^{(\texttt{5})}$ \\ 
        $\mathbf{S}_{14}$ & 0.71 & 0.77 & 0.44 & 0.35 & 0.46 & 0.44 & 0.64 & 0.14 & 0.39 & 0.10 & 0.61$^{(\texttt{4})}$ & 0.65$^{(\texttt{6})}$ & 0.63$^{(\texttt{4})}$ & 0.53$^{(\texttt{7})}$ & 0.58$^{(\texttt{7})}$ & 0.55$^{(\texttt{7})}$ \\ 
        $\mathbf{S}_{15}$ & 0.90 & 0.46 & 0.76 & 0.45 & 0.28 & 0.76 & 0.71 & 0.18 & 0.50 & 0.10 & 0.66$^{(\texttt{3})}$ & 0.73$^{(\texttt{3})}$ & 0.68$^{(\texttt{3})}$ & 0.68$^{(\texttt{3})}$ & 0.73$^{(\texttt{4})}$ & 0.69$^{(\texttt{3})}$ \\ 
	\bottomrule
    \end{tabular}%
\label{tab:exemplary-alternatives}
\end{table*}%

The visualizations in Figure~\ref{fig:sandcastle3-weightcastle3-05-06-1-I-A-T} confront the shapes of MSD-space and WMSD-space. The latter can be regarded as a natural generalization of the MSD-space that incorporates different weights assigned to the criteria. For $\mathbf{w} \neq \mathbf{1}$, the WMSD-space is characterized by a potentially larger number of vertices and a possibly smaller range of mean values than the MSD-space. 
Naturally, preference information given by the decision maker in the form of weights influences not only the shape of the space but also the position of the alternatives within those spaces. For example, the positions of $\mathbf{S}_{8}$ and $\mathbf{S}_{9}$ change quite drastically, as the two alternatives almost swap their positions: they are both characterized by a very similar $std(\mathbf{u})$ and $std^{01}_\mathbf{w}(\mathbf{v})$, but $mean(\mathbf{S}_{8})=0.68>mean(\mathbf{S}_{9})=0.57$ whereas 
$mean^{01}_\mathbf{w}(\mathbf{S}_{8})=0.35<mean^{01}_\mathbf{w}(\mathbf{S}_{9})=0.50$.
As expected, the images of the best possible alternative ($\mathbf{1}$) and the worst possible one ($\mathbf{0}$) have fixed relative positions no matter what the values of weights are, in the sense that they are always situated, respectively, in the rightmost and leftmost vertices of MSD-space and WMSD-space (see $\mathbf{S}_{12}$ and $\mathbf{S}_{10}$ in Figure~\ref{fig:sandcastle3-weightcastle3-05-06-1-I-A-T}).

The change of alternative position imposed by the incorporation of weights can also be observed for alternatives that are characterized by different vectors in $\mathit{US}$, but by a single point in MSD-space. To illustrate this case, let us consider $\mathbf{S}_{6}$ and $\mathbf{S}_{14}$. As shown in Table~\ref{tab:exemplary-alternatives} and Figure~\ref{fig:sandcastle3-weightcastle3-05-06-1-I-A-T} vectors $[0.59, 0.49, 0.82]$ (the $\mathit{US}$ image of $\mathbf{S}_{6}$) and $[0.71, 0.77, 0.44]$ (the $\mathit{US}$ image of $\mathbf{S}_{14}$) are characterized by $mean(\mathbf{S_{6}}) = mean(\mathbf{S_{14}}) = 0.64$ and $std(\mathbf{S_{6}}) = std(\mathbf{S_{14}}) = 0.14$.
As a result, those vectors share the same point in MSD-space and are thus identically evaluated by the aggregations $\mathsf{I}(\mathbf{u})$,  $\mathsf{A}(\mathbf{u})$ and $\mathsf{R}(\mathbf{u})$.
However, the two alternatives do not share the same point in WMSD-space, as their positions in WMSD-space are influenced by the weights. The weights affect the vectors in $\mathit{US}$, which are different for $\mathbf{S}_{6}$ and $\mathbf{S}_{14}$. As a result, the $mean^{01}_\mathbf{w}(\mathbf{S}_{6})=0.50 \neq mean^{01}_\mathbf{w}(\mathbf{S}_{14})=0.39$, which implies that the two alternatives are not identically evaluated by the considered aggregations (even though $std^{01}_\mathbf{w}(\mathbf{S}_{6}) = std^{01}_\mathbf{w}(\mathbf{S}_{14})=0.10$). In particular, $\mathbf{S}_{6} \succ_{\mathsf{I_{w}}(\mathbf{v}), \mathsf{A_{w}}(\mathbf{v}), \mathsf{R_{w}}(\mathbf{v})} \mathbf{S}_{14}$ (see Table~\ref{tab:exemplary-alternatives}).
Similarly, alternatives that are characterized by the same point in WMSD-space do not share, in general, the same point in MSD-space. In particular, $\mathbf{S}_{6}$ and $\mathbf{S}_{15}$ are characterized by different vectors in $\mathit{US}$, namely $[0.59, 0.49, 0.82]$ for $\mathbf{S}_{6}$ and $[0.90, 0.46, 0.76]$ for $\mathbf{S}_{15}$, but are characterized by $mean^{01}_\mathbf{w}(\mathbf{S}_{6}) = mean^{01}_\mathbf{w}(\mathbf{S}_{15}) = 0.50$ and $std^{01}_\mathbf{w}(\mathbf{S}_{6}) = std^{01}_\mathbf{w}(\mathbf{S}_{15}) = 0.10$, which puts them in the very same position in WMSD-space. This however, does not imply sharing the same point in MSD-space, since this position is influenced by the vector of weights. Consequently, $\mathbf{S}_{6}$ and $\mathbf{S}_{15}$ are evaluated identically by $\mathsf{I_{w}}(\mathbf{v})$, $\mathsf{A_{w}}(\mathbf{v})$ and $\mathsf{R_{w}}(\mathbf{v})$, but differently by $\mathsf{I}(\mathbf{u})$, $\mathsf{A}(\mathbf{u})$ and $\mathsf{R}(\mathbf{u})$ (Figure~\ref{fig:sandcastle3-weightcastle3-05-06-1-I-A-T}).

Although the MSD-space and WMSD-space share the same character of the isolines under particular aggregation, the change of the alternatives' position across those spaces caused by the weights influences the final ratings of the alternatives.
Let us look again at $\mathbf{S}_{8}$ and $\mathbf{S}_{9}$. It is clear that because of weights, their rankings are reversed: $\mathbf{S}_{8} \succ_{\mathsf{I}(\mathbf{u}), \mathsf{A}(\mathbf{u}),\mathsf{R}(\mathbf{u}) } \mathbf{S}_{9}$ versus $\mathbf{S}_{9} \succ_{\mathsf{I_{w}}(\mathbf{v}), \mathsf{A_{w}}(\mathbf{v}), \mathsf{R_{w}}(\mathbf{v})} \mathbf{S}_{8}$ (Table~\ref{tab:exemplary-alternatives}). Alternative $\mathbf{S}_{8}$ is much better than $\mathbf{S}_{9}$ on the first two criteria, but their importance was diminished by the weights being 0.5 and 0.6, causing $\mathbf{S}_{9}$ to climb higher in those rankings that take weights into account.

The above analysis compared the MSD-space and WMSD-space from the viewpoint of criteria weights. Let us now illustrate some trade-offs between the values of $mean^{01}_\mathbf{w}(\mathbf{v})$ and $std^{01}_\mathbf{w}(\mathbf{v})$ in WMSD-space, and show how they influence the final rankings under different aggregations.
The interplay of $mean^{01}_\mathbf{w}(\mathbf{v})$ and $std^{01}_\mathbf{w}(\mathbf{v})$ in the context of preferences, as formalized in Table~\ref{tab:preference-related-interplay}, is illustrated in the WMSD-space by the color reflecting the aggregation value imposed for each point of the space. Getting a higher ranking position requires an increase in the aggregation value, which is reflected by a change of the alternative's color towards dark red. This can naturally be achieved when the alternative obtains more desirable values on the criteria. In our example from Table~\ref{tab:students}, this means that a student should get better marks in some subjects while not worsening them in any other subject. This would result in the increase of $mean^{01}_\mathbf{w}(\mathbf{v})$, which can, however, be hard to achieve, or in some cases even impossible. However, the preference related interplay in the WMSD-space between $mean^{01}_\mathbf{w}(\mathbf{v})$ and $std^{01}_\mathbf{w}(\mathbf{v})$ shows 
other ways to influence the ranking even without increasing $mean^{01}_\mathbf{w}(\mathbf{v})$. 

First, let us focus on three alternatives characterized by the same $mean^{01}_\mathbf{w}(\mathbf{v})=0.5$: $\mathbf{S}_{3}$, $\mathbf{S}_{6}$ and $\mathbf{S}_{9}$ (analogous discussion is valid for, e.g., $\mathbf{S}_{1}$, $\mathbf{S}_{4}$ and $\mathbf{S}_{7}$). The ranking under the $\mathsf{I_{w}}(\mathbf{v})$ aggregation is the following: $\mathbf{S}_{3} \succ_{\mathsf{I_{w}}(\mathbf{v})} \mathbf{S}_{6} \succ_{\mathsf{I_{w}}(\mathbf{v})} \mathbf{S}_{9}$, as opposed to $\mathsf{A_{w}}(\mathbf{v})$, where $\mathbf{S}_{9} \succ_{\mathsf{A_{w}}(\mathbf{v})} \mathbf{S}_{6} \succ_{\mathsf{A_{w}}(\mathbf{v})} \mathbf{S}_{3}$. Clearly, a change in the $std^{01}_\mathbf{w}(\mathbf{v})$, with no change of $mean^{01}_\mathbf{w}(\mathbf{v})$, is enough to influence the rankings. Under aggregation $\mathsf{I_{w}}(\mathbf{v})$ less variant values of the criteria are preferred, as $std^{01}_\mathbf{w}(\mathbf{v})$ is of cost-type for this aggregation. In contrast, under the $\mathsf{A_{w}}(\mathbf{v})$ aggregation an increase of $std^{01}_\mathbf{w}(\mathbf{v})$ results in the increase of the aggregation function. Aggregation $\mathsf{R_{w}}(\mathbf{v})$ on the other hand, resembles aggregation $\mathsf{A_{w}}(\mathbf{v})$ when $mean^{01}_\mathbf{w}(\mathbf{v})< \frac{mean(\mathbf{w})}{2}$ and aggregation $\mathsf{I_{w}}(\mathbf{v})$ when $mean^{01}_\mathbf{w}(\mathbf{v})> \frac{mean(\mathbf{w})}{2}$. In the very middle of WMSD-space, i.e. when $mean^{01}_\mathbf{w}(\mathbf{v})=\frac{mean(\mathbf{w})}{2}$, the change of $std^{01}_\mathbf{w}(\mathbf{v})$ has no effect on the ranking at all. 
This brings us to the conclusion that the isolines in the WMSD-space can guide the decision maker as to what actions need to be taken in order to influence the ranking without changing the $mean^{01}_\mathbf{w}(\mathbf{v})$ of the alternative. 

Now, let us focus on another set of alternatives: $\mathbf{S}_{4}$, $\mathbf{S}_{5}$ and $\mathbf{S}_{6}$ (analogous discussion is valid for, e.g., $\mathbf{S}_{10}$, $\mathbf{S}_{1}$, $\mathbf{S}_{2}$, $\mathbf{S}_{3}$ and $\mathbf{S}_{12}$). They are characterized by the same $std^{01}_\mathbf{w}(\mathbf{v})$, but different $mean^{01}_\mathbf{w}(\mathbf{v})$. Under all the three considered aggregations, the alternatives are ranked the same: 
$\mathbf{S}_{6} \succ_{\mathsf{I_{w}}(\mathbf{v}), \mathsf{A_{w}}(\mathbf{v}), \mathsf{R_{w}}(\mathbf{v})} \mathbf{S}_{5} \succ_{\mathsf{I_{w}}(\mathbf{v}), \mathsf{A_{w}}(\mathbf{v}), \mathsf{R_{w}}(\mathbf{v})} \mathbf{S}_{4}$. This results from the fact that $mean^{01}_\mathbf{w}(\mathbf{v})$ is of type gain for all the considered aggregations. Thus, moving to the right in the WMSD-space (i.e., keeping the same $std^{01}_\mathbf{w}(\mathbf{v})$ and only increasing $mean^{01}_\mathbf{w}(\mathbf{v})$) always increases the aggregation functions.

Last but not least, the alternative's rating can be influenced by a simultaneous change in $mean^{01}_\mathbf{w}(\mathbf{v})$ and $std^{01}_\mathbf{w}(\mathbf{v})$.
A rather hard to predict compensation of those two is clearly visible in the WMSD-space, as the change in rating is equivalent to `switching between' isolines. Let us consider $\mathbf{S}_{9}$ and $\mathbf{S}_{13}$ (analogous discussion is valid for $\mathbf{S}_{7}$ and $\mathbf{S}_{11}$).
Observe that the rankings of those two alternatives are different under different aggregations: $\mathbf{S}_{9} \succ_{\mathsf{I_{w}}(\mathbf{v})} \mathbf{S}_{13}$, but $\mathbf{S}_{13} \succ_{\mathsf{R_{w}}(\mathbf{v})} \mathbf{S}_{9}$. This change results from the fact that the isolines for the $\mathsf{R_{w}}(\mathbf{v})$ aggregation `straighten up' while moving towards the middle values of $mean^{01}_\mathbf{w}(\mathbf{v})$, while the isolines for the $\mathsf{I_{w}}(\mathbf{v})$ aggregation keep the same concentric character. The isolines are thus a visual representation of the trade-offs between 
$mean^{01}_\mathbf{w}(\mathbf{v})$ and $std^{01}_\mathbf{w}(\mathbf{v})$ for different aggregations.

\subsection{Index of Economic Freedom}
\label{sec:freedom}
The second case study is based on publicly available data from the Index of Economic Freedom~\cite{Index_Economic_Freedom_url}, which covers 12 freedoms---from property rights to tax burdens---in 184 countries. The data has been annually collected for almost 30 years now by The Heritage Foundation~\cite{Index_Ecom_Freedom_23} and served as the basis for many case studies and analyses~\cite{Lima_Silva_23,Balkans_23,Dinc_22,Lima_Silva_20,Brkic_20}. 
In particular, our case study is based on the data gathered for the $25^{th}$ anniversary of the Index in 2019, which were used by \citet{Lima_Silva_20}.

Economic freedom is understood as the right of every human to control their own labor and property. 
Within the Index, 12 factors are measured and grouped into four categories:
Rule of Law, Government Size, Regulatory Efficiency, Open Markets. There are three factors per category, each factor is graded on a 0--100 scale of type gain. Details on how the values of factors are determined for the considered countries are available in~\cite{Index_Ecom_Freedom_19}.
For the purpose of this case study, we have limited the Index only to the 12 countries of South America and aggregated the criteria by taking the mean of factors forming each category (Table~\ref{tab:SA_data}). To ensure reproducibility, the raw data from the Heritage Foundation, their transformations, and final rankings under different aggregations and weights are available in the online supplementary materials.

\begin{table}[htbp]
    \centering
  \caption{Descriptions of alternatives for the second case study; South American countries described by four criteria of type gain: Rule of Law, Government Size, Regulatory Efficiency, Open Markets. Each criterion is formed as a mean of three factors obtained from the Index of Economic Freedom~\cite{Index_Economic_Freedom_url}.
  }
  \label{tab:SA_data}
    \begin{tabular}{llcccc}
    \toprule
    {ID}  & {Country} & Rule of Law & Gov. Size & Reg. Eff. & Open Markets\\
    \midrule
    ARG	&	Argentina	&	41.93	&	50.60	&	54.50	&	61.67	\\
    BOL	&	Bolivia		&	17.50	&	49.77	&	60.17	&	41.80	\\
    BRA	&	Brazil		&	45.70	&	43.87	&	61.77	&	56.33	\\
    CHL	&	Chile		&	62.43	&	82.43	&	75.37	&	81.27	\\
    COL	&	Colombia	&	42.33	&	76.17	&	75.17	&	75.33	\\
    ECU	&	Ecuador		&	27.13	&	54.87	&	58.60	&	47.13	\\
    GUY	&	Guyana		&	39.27	&	71.33	&	66.07	&	50.60	\\
    PRY	&	Paraguay	&	31.67	&	90.50	&	54.50	&	70.53	\\
    PER	&	Peru		&	40.63	&	85.07	&	71.73	&	73.80	\\
    SUR	&	Suriname	&	35.60	&	52.57	&	59.27	&	44.87	\\
    URY	&	Uruguay		&	65.47	&	71.53	&	73.03	&	64.53	\\
    VEN	&	Venezuela	&	09.53	&	50.13	&	20.63	&	23.33	\\
    \bottomrule
    \end{tabular}
\end{table}

The case study focuses on the $\mathsf{R_{\mathbf{w}}}(\mathbf{v})$ aggregation, since it is the one most commonly used in practice. The following four sets of weights are considered as examples of preference information given by different decision-makers:
\begin{itemize}
    \item $\mathbf{w_{1}} = [1.00, 1.00, 1.00, 1.00]$,
    \item $\mathbf{w_{2}} = [0.25, 1.00, 0.25, 0.50]$,
    \item $\mathbf{w_{3}} = [0.50, 1.00, 0.25, 0.25]$, 
    \item $\mathbf{w_{4}} = [1.00, 0.66, 0.33, 0.00]$.
\end{itemize}
These sets of weights were not chosen with any domain-specific rationale but to demonstrate different situations arising when working with weights. In particular, the first weight vector expresses equal importance of all criteria, allowing for the visualization of the alternatives in MSD-space, as opposed to the other sets of weights, which require WMSD-space. The second and third vectors were chosen to be permutations of each other to illustrate that the order of weights does not affect the shape of WMSD-space. Finally, the last weight vector completely eliminates the influence of the Open Markets criterion on the country rankings by setting its weight to zero; this makes $n_p = 3 < n = 4$.

\begin{figure*}[h!]
\centering
\includegraphics[width=0.97\textwidth]{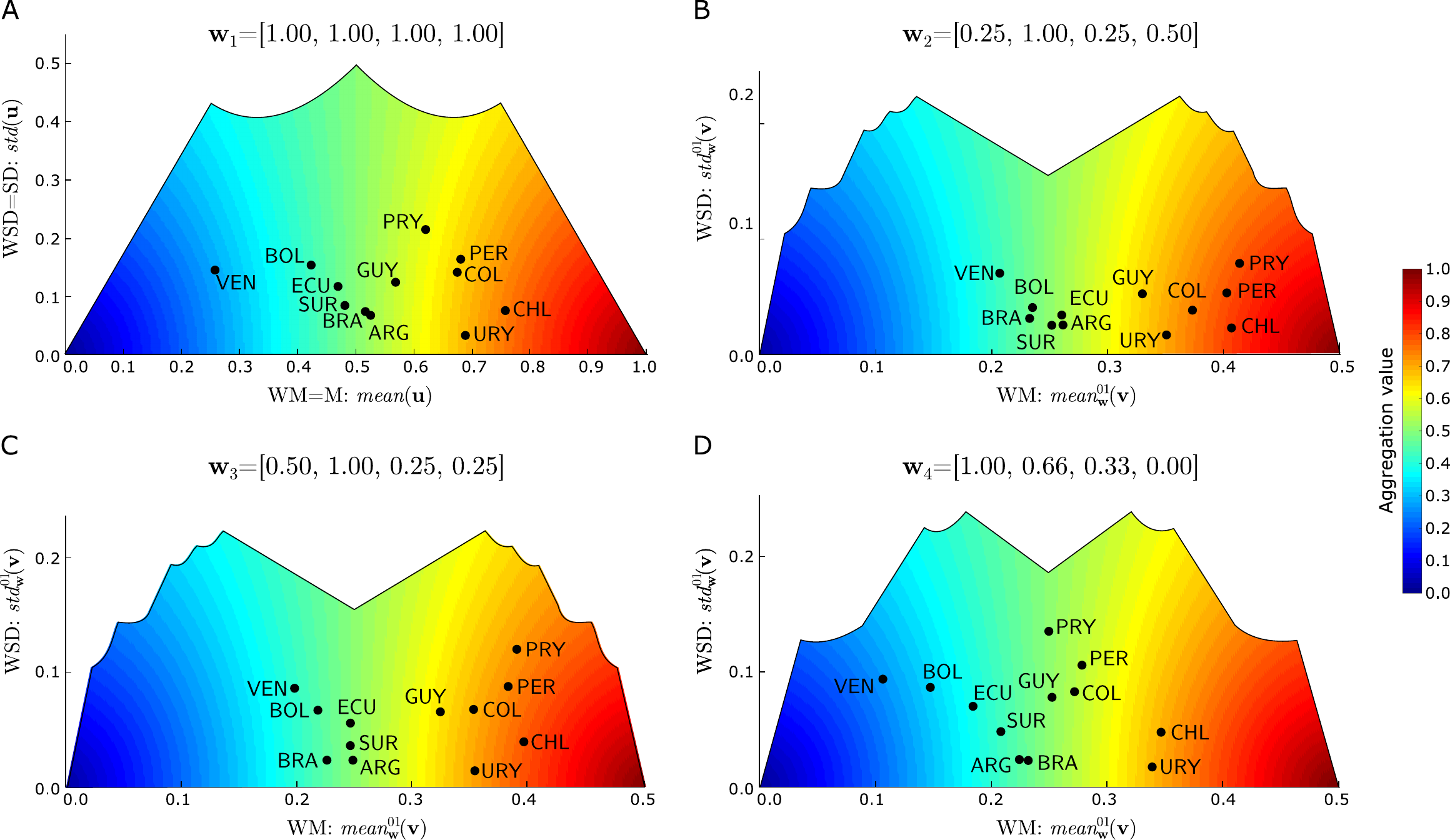}
\caption{The visualization of preference as expressed by $\mathsf{R_{w}}(\mathbf{v})$ aggregation in WMSD-spaces defined for four different weight vectors. The color map reflects the preference: dark blue---the least preferred, dark red---the most preferred. Note that the sub-figures are on different scales; we have scaled up the WMSD sub-figures to increase their readability.}
\label{fig:freedom}
\end{figure*}

Figure~\ref{fig:freedom} shows the countries of South America superimposed on a gradient depicting the preference expressed by $\mathsf{R_{\mathbf{w}}}(\mathbf{v})$ in WMSD-space. The final country rankings are gathered in Table~\ref{tab:SA_R_rankings}.

\begin{table}[htbp]
  \centering
  \caption{Rankings of South American countries resulting from $\mathsf{R_{\mathbf{w}}}(\mathbf{v})$ aggregation under different weight vectors.}
    \begin{tabular}{rcccccccc}
    \toprule
    \multirow{2}{*}{Rank} &
    \multicolumn{2}{c}{$\mathbf{w_{1}}$}  & 
    \multicolumn{2}{c}{$\mathbf{w_{2}}$}  & 
    \multicolumn{2}{c}{$\mathbf{w_{3}}$}  & 
    \multicolumn{2}{c}{$\mathbf{w_{4}}$}  \\
    & \multicolumn{2}{c}{$[1.00, 1.00, 1.00, 1.00]$} & 
    \multicolumn{2}{c}{$[0.25, 1.00, 0.25, 0.50]$} & 
    \multicolumn{2}{c}{$[0.50, 1.00, 0.25, 0.25]$} &
    \multicolumn{2}{c}{$[1.00, 0.66, 0.33, 0.00]$} \\
    \midrule
   1 & CHL   & 0.746 & CHL   & 0.806 & CHL   & 0.775 & CHL   & 0.684 \\
   2 & URY   & 0.685 & PER   & 0.787 & PER   & 0.725 & URY   & 0.677 \\
   3 & PER   & 0.659 & PRY   & 0.785 & PRY   & 0.713 & PER   & 0.548 \\
   4 & COL   & 0.658 & COL   & 0.738 & URY   & 0.701 & COL   & 0.539 \\
   5 & PRY   & 0.599 & URY   & 0.700 & COL   & 0.685 & GUY   & 0.503 \\
   6 & GUY   & 0.564 & GUY   & 0.652 & GUY   & 0.634 & PRY   & 0.501 \\
   7 & ARG   & 0.521 & ARG   & 0.524 & ARG   & 0.497 & BRA   & 0.464 \\
   8 & BRA   & 0.519 & ECU   & 0.523 & ECU   & 0.497 & ARG   & 0.453 \\
   9 & SUR   & 0.481 & SUR   & 0.507 & SUR   & 0.494 & SUR   & 0.424 \\
  10 & ECU   & 0.471 & BOL   & 0.474 & BRA   & 0.456 & ECU   & 0.382 \\
  11 & BOL   & 0.430 & BRA   & 0.471 & BOL   & 0.444 & BOL   & 0.321 \\
  12 & VEN   & 0.283 & VEN   & 0.426 & VEN   & 0.412 & VEN   & 0.262 \\
    \bottomrule
    \end{tabular}%
  \label{tab:SA_R_rankings}%
\end{table}%

As can be noticed in Figure~\ref{fig:freedom}, the weights have a clear influence both on the shape of WMSD-spaces and the rankings of the alternatives.
First, observe that $mean(\mathbf{w_{2}})=mean(\mathbf{w_{3}})=mean(\mathbf{w_{4}})=0.5$, resulting in the same WM range (x-axis) in Figures~\ref{fig:freedom}B, C and D. Additionally, vector $\mathbf{w_{3}}$ is simply a permutation of $\mathbf{w_{2}}$, thus the whole shape of the respective WMSD-spaces is exactly the same (see Figure~\ref{fig:freedom}B and C). Nonetheless, the positions of the alternatives within those shapes differ, leading to different rankings. 
The weights also naturally influence the number of vertices in the WMSD-spaces, which is, in particular, reflected by a smaller number of vertices when some criteria are given weights equal to zero (compare Figure~\ref{fig:freedom}B and~D).

Looking at the rankings under different weight vectors (Table~\ref{tab:SA_R_rankings}), one notices the inevitable changes in country ratings imposed by incorporating weights. For example, Uruguay shifts from the second position under $\mathbf{w_{1}}$ or $\mathbf{w_{4}}$ to as far as the fifth position under $\mathbf{w_{2}}$. 
Interestingly, for particular weight vectors, some countries are almost indiscernible as their values of the $\mathsf{R_{\mathbf{w}}}(\mathbf{v})$ aggregation differ only slightly, e.g., Argentina, Ecuador, and Suriname under $\mathbf{w_{3}}$. The WMSD visualizations (Figure~\ref{fig:freedom}) explain in terms of means and standard deviations why the ranking positions of those countries are (almost) the same. Observe that the considered countries are located in the green region, i.e., very close to the middle of the WM range (x-axis), which happens to be the region where $std^{01}_\mathbf{w}(\mathbf{v})$ hardly influences the rankings (Table\ref{tab:preference-related-interplay}). Thus, despite occupying different points in the WMSD-space, the countries are ranked almost equally. 

\begin{figure*}[h!]
\centering
\includegraphics[width=0.48\textwidth]{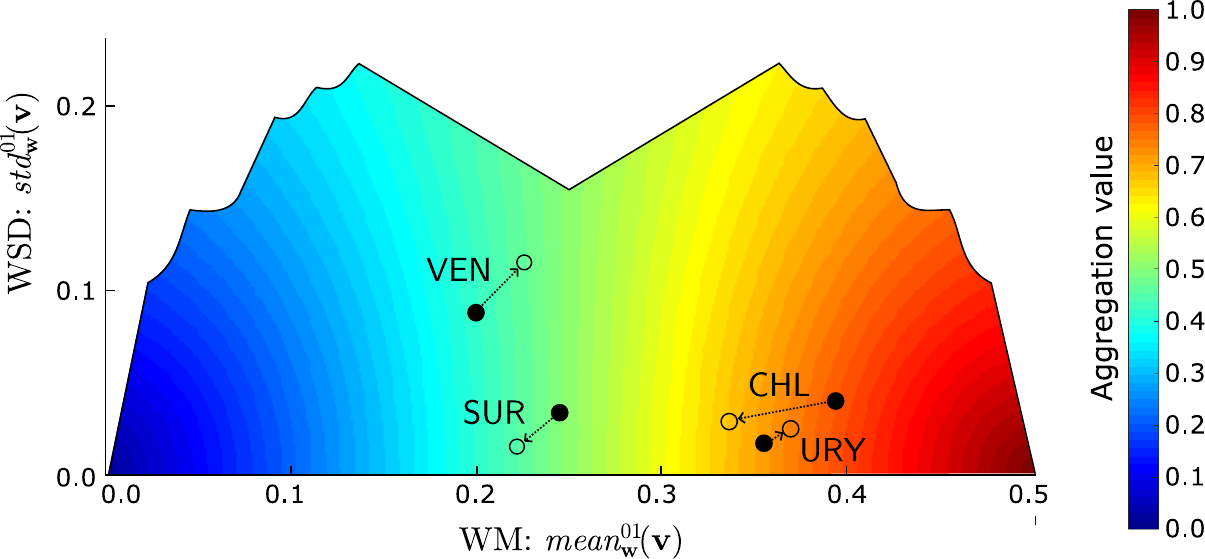}
\caption{The visualization in WMSD-space of the change of the countries' positions between 2019 (solid circles) and 2023 (empty circles) under $\mathsf{R_{w}}(\mathbf{v})$ aggregation and weight vector $\mathbf{w_{3}}=[0.50, 1.00, 0.25, 0.25]$. }
\label{fig:2019_2023_comparison}
\end{figure*}

Since the Index of Economic Freedom is updated annually, it would also be interesting to visually compare the data from 2019 and 2023. This can be easily done using WMSD-space. In Figure~\ref{fig:2019_2023_comparison}, we show such a comparison for four countries countries: Chile, Uruguay, Suriname, and Venezuela.
Observe that under a particular aggregation ($\mathsf{R_{\mathbf{w}}}(\mathbf{v})$) and weight vector ($\mathbf{w_{3}}=[0.50, 1.00, 0.25, 0.25]$) the shape of the WMSD-space and the isolines of the aggregation function are fixed. Thus, the comparison of the data from various years only requires superimposing that data on the WMSD-space. 
The countries' positions in the year 2019 are depicted by solid circles, whereas empty circles mark the countries' positions in 2023.
This dynamic perspective shows that Venezuela's and Uruguay's positions improved while Suriname's and Chile's got worse. As a result, in 2023, Venezuela outranks Suriname and Uruguay outranks Chile, which was not the case in 2019. All of the considered ranking transitions resulted from the changes on both $mean^{01}_\mathbf{w}(\mathbf{v})$ and $std^{01}_\mathbf{w}(\mathbf{v})$. 
A closer inspection reveals that out of the four considered countries, Chile is characterized by the biggest decrease in $mean^{01}_\mathbf{w}(\mathbf{v})$, which directly caused Chile's drop in the ranking. On the other hand, the description of Venezuela in terms of the analyzed criteria became notably diversified, which was reflected by the biggest change in $std^{01}_\mathbf{w}(\mathbf{v})$. Since Venezuela is situated in the left-hand side of the WMSD-space ($mean^{01}_\mathbf{w}(\mathbf{VEN})<\frac{mean(\mathbf{w_{3}})}{2}=0.25$) such an increase of variety had a positive effect on the country's ranking position.   
The above considerations show that WMSD-space is a useful tool not only for visualizing the impact that weight and aggregations have on the final rankings but also for analyzing changes in rankings over time.  

\begin{figure*}[h!]
\centering
\includegraphics[width=0.48\textwidth]{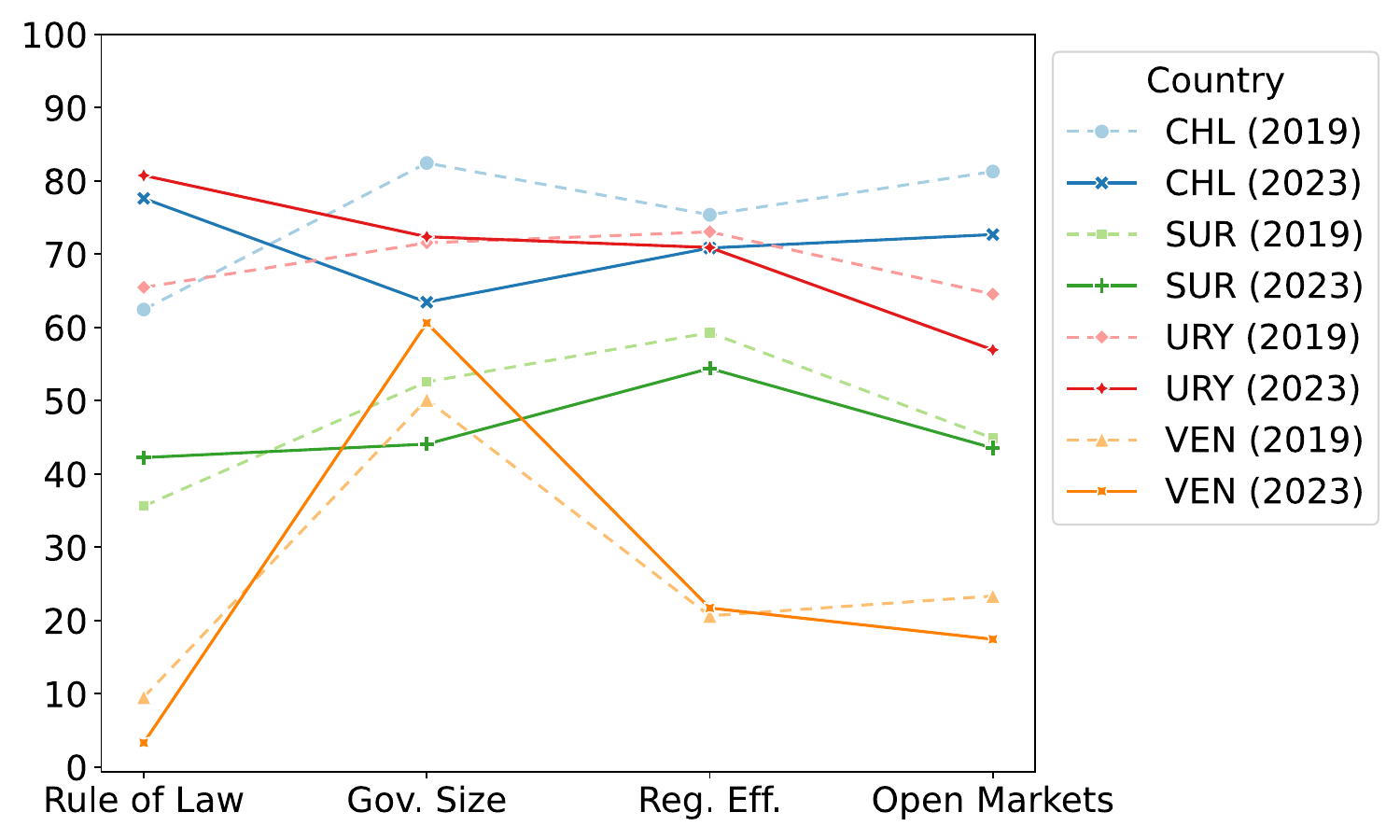}
\caption{Parallel coordinates plots comparing the countries' criteria values in 2019 (dashed line) and 2023 (solid line). }
\label{fig:parallel_coordinates}
\end{figure*}

As mentioned in Section~\ref{sec:introduction}, currently available methods that visualize dependencies between alternatives and their rankings are very limited. The most common approach to visualizing alternatives is the use of parallel coordinates plots~\cite{ZPN_23}. Figure~\ref{fig:parallel_coordinates} presents such a plot depicting the criteria values of Chile, Uruguay, Suriname, and Venezuela in the 2019 and 2023 economic freedom rankings. The parallel coordinates plot is capable of showing all the criteria values for each alternative. With only four criteria and eight alternatives, this makes Figure~\ref{fig:parallel_coordinates} fairly readable. However, with more criteria and alternatives, this type of visualization would very quickly become cluttered, whereas WMSD-space can visualize any number of criteria and tens of alternatives and remain readable. Moreover, the parallel coordinates plot does not take into account criteria weights and does not clearly show the ranking. For example, it is practically impossible to notice that Venezuela is ranked higher than Suriname in 2023. Similarly, the change in ranking for Chile and Uruguay is hard to spot in the parallel coordinates plot, whereas it is clearly visible on the WMSD-space visualization (Figure~\ref{fig:2019_2023_comparison}). Finally, the gradient background of the WMSD visualization shows the properties of the used aggregation function, whereas the parallel coordinates are completely decoupled from the used aggregation.

Summing up, the main goal of the economic freedom case study was to show that WMSD-space visualizations can be applied to real-world TOPSIS ranking problems. Since most publications on TOPSIS are applications of the method rather than contributions to the methodology, it was important for us to provide practitioners with a problem setting they are familiar with. That being said, the economic freedom case study highlights several general decision-making implications connected with TOPSIS: 
\begin{itemize}
    \item Criteria weights affect the possible distances to the ideal and anti-ideal solution and, hence, the possible aggregation function values and the shape of WMSD-space.
    \item Criteria weights affect aggregation values of the alternatives and, hence, their ranking positions.
    \item A permutation of weights (the same set of weights applied to different criteria) does not change the space of possible ideal/anti-ideal distances and, hence, does not change WMSD-space (although it can change ranking positions of individual alternatives). 
    \item Criteria weights do not affect the properties of aggregation functions and, hence, their gradient in WMSD-space. 
\end{itemize}

In the next section, we will discuss the limitations of WMSD visualizations and the potential for their further modifications and applications.
  
\section{Discussion}
\label{sec:discussion}
WMSD-space offers a visualization method for the standard TOPSIS methodology. We have shown that multicriteria alternative ranking problems can be visualized in 2D in terms of means and standard deviations, even when the number of criteria exceeds two and even when the criteria are weighted. Since means and standard deviations used in the WMSD-space are easily interpretable, the space offers an explainable view of TOPSIS rankings, showing whether the means or standard deviations of particular alternatives impacted a given alternative's ranking position. Moreover, by showing the aggregation values of all possible alternatives as a colored gradient, WMSD-space shows the properties of TOPSIS aggregation functions.

Apart from restating the main characteristic of the proposed visualization method, it is worth listing the potential limitations of WMSD-space. First, we note that since WMSD-space is a visualization tool customized for TOPSIS, it inherits the limitations of the overall TOPSIS methodology. This implies requiring precise input data (without any ambiguity), batch-like processing mode (without any user interaction/parametrization), and a linear pre-order output result (without the relation of incomparability). Moreover, a distance measure for alternative representations must be defined, which often requires the criteria defined using finite ranges. Finally, the considered weights must be non-negative, which is an accepted practice---in this common interpretation, weights express the magnitude of criterion importance and, as such, cannot be negative.
In addition to limitations inherited from TOPSIS, there occurs a limitation of the visualization method itself, namely the need to use the Euclidean distance measure. The dimensionality reduction in WMSD-space is made possible because of the application of the Pythagorean theorem, which is derived from the axioms of Euclidean geometry. Thus, for the visualization to be possible regardless of the number of criteria, the Euclidean distance is needed. Moreover, as far as weighting is concerned, we assume the standard TOPSIS methodology. As a result, weights are applied multiplicatively, which thus constitutes a linear operation. Potential attempts to make TOPSIS use weights that are in a non-linear relationship with criteria values would require an explicit modification of the method itself. Additionally, if our WMSD-space based visualization were to handle this modified version of TOPSIS, it would require new, potentially non-linear versions of projection and rejection. This is because a non-linear transformation from $US$ to $VS$ would, in particular, result in a non-linear diagonal $D_{\mathbf{0}}^{\mathbf{w}}$. Should the results of projections onto and rejections from such a non-linear diagonal be possible and unique, the visualization methodology presented in this paper would still hold.

It is also worth noting that the proposed WMSD-space visualization does not introduce any significant computational overhead compared to the standard TOPSIS procedure. More precisely, to calculate the coordinates of alternatives in WMSD-space, one needs to calculate the weight-scaled means and standard deviations of each alternative. For $m$ alternatives described by $n$ criteria, this requires $O(mn)$ time.

Finally, an issue worth discussing is the potential for using WMSD-space to design improvement actions. As discussed in Section~\ref{sec:case} (Figure~\ref{fig:2019_2023_comparison}), WMSD-space can show what actions (in terms of changing the mean or standard deviation of an alternative) can be taken to change the alternative's ranking position. Of course, such mean and standard deviation actions can be then translated into actions on criteria values. In this context, the IA-WMSD property (Table~\ref{tab:preference-related-interplay}) sums up the relations between mean and standard deviation for different aggregations. It is, therefore, easy to foresee in WMSD-space whether the increase or decrease of mean or standard deviation will lead to the increase of the value of the aggregation function. As such, WMSD-space visualizations could also prove useful in the potential sensitivity analyses of TOPSIS, if such were attempted. Even more importantly, TOPSIS is often applied in high-stake domains that impact societies. As an example, in a recent sustainable energy planning project~\cite{bottero} environment, economy, technical, and social experts each provided their own preference in the form of weights on criteria concerning the costs and impact of using renewable energy sources on a university campus. Such use cases could benefit from WMSD-space visualizations, which could be used to compare rankings obtained from weights provided by different experts. Importantly, WMSD-space can help interpret how the mean and standard deviation of an alternative affect its position in a given ranking. Therefore, as future work, our goal is to make WMSD-space visualizations easily available through an open-source tool and to apply them to real-world decision-making problems.
 
\section{Conclusions}
\label{sec:conclusions}
In this paper, we have put forward a visual-based method for explaining TOPSIS rankings in practical decision support applications with expert-defined criteria weights. To this end, weight-scaled means and standard deviations of alternatives were defined as generalizations of means
and standard deviations. Formalizing their relationship with distances of an alternative to predefined ideal/anti-ideal points (IA-WMSD property), we have proposed a generalization of MSD-space called WMSD-space.
Unlike MSD-space, which is based on regular means and standard deviation, WMSD-space is based on weight-scaled means and standard deviations of alternatives. Both, however, are equally capable of representing alternatives and aggregation functions in a plane regardless of the number of considered criteria. What differentiates WMSD from MSD is taking into account the criteria weights. As such, WMSD-space is a tool for visual-based comparisons of different aggregation functions and the impact that weights defined by experts have on the final rankings. 
Therefore, the answer to the question asked in the Introduction, namely: \textit{Is it possible to propose a visualization method that generalizes MSD-space to weighted criteria?}, is thus definitely positive.

We stress that our paper proposes a visualization technique for the standard TOPSIS method, not a new version of TOPSIS. The novelty comes from considering all possible alternative representations (instead of particular datasets), which makes it possible to visualize gradients of aggregation functions, and the fact that our visualization method is always two-dimensional, even when the number of criteria is greater than two. 
To highlight the practical usefulness of the proposed visualization, two case studies were conducted on a dataset of students described in terms of school grades and on a dataset of countries described in terms of factors constituting the Index of Economic Freedom.
Using WMSD-space visualizations, we discuss how weights affect rankings of alternatives under various TOPSIS aggregations and compare
the effects of weights provided by multiple experts.

As future research, the WMSD-space methodology could be extended to take into account uncertainty connected with the input data or criteria weights. This could involve representing data in the form of fuzzy sets, as they are commonly used to model the imprecision and uncertainty of data in many application domains~\cite{CHE00,Lin23ASC,bottani2006fuzzy,aydougdu2023complex}. We expect that such fuzzy input will result in alternatives represented not as points but as polygons---subregions of WMSD-space. On the other hand, the uncertainty of the decision-makers, represented as fuzzy criteria weights, would require a 3D visualization that would allow for interpolation between WMSD spaces of potentially different shapes.
Future research could also include the development of TOPSIS modifications that would control the impact that alternative means and standard deviations have on the final rankings. Such modifications would involve a parameter letting the user define whether the rankings should be more influenced by the weight-scaled means or standard deviations. Such an analysis would generalize the results presented here to other distance-based MCDA ranking methods such as UTA~\cite{LAGSIS82} or SAW~\cite{ciardiello2023comparison}.
Similarly, WMSD visualization may be potentially applied to TOPSIS generalizations based on three-way decision~\cite{wang2022bmw,zhan2023modified}.
Finally, to meet the needs of practitioners, we plan to develop a publicly available open-source interactive dashboard for WMSD visualizations of user-provided datasets. Combined with a new spectrum of aggregations and possible improvement actions, it would make a valuable tool for hands-on multi-criteria decision analysis.

\section*{Acknowledgments}
\noindent
We would like to thank Adam Ciesiółka, Dariusz Grynia, Bogna Kalinowska, and Maciej Woś for their work on the Python implementation of the WMSD procedure. This research was partly funded by the National Science Centre, Poland, grant number: 2022/47/D/ST6/01770. For the purpose of Open Access, the author has applied a CC-BY public copyright license to any Author Accepted Manuscript (AAM) version arising from this submission.

\bibliographystyle{elsarticle-num-names}
\bibliography{MSD-space}

\newpage

\appendix

\setcounter{figure}{0}
\renewcommand{\thefigure}{A\arabic{figure}}

\section{Detailed Example: Space Transformations, Distance Computations, IA-WMSD Property Verifications and Aggregation Values}
\label{app:detailed_example}


\noindent
All fractional values in the following computations are presented with four significant digits. For a summary of the notation used, see Table~\ref{tab:notation}.

\subsection{A $2$-dimensional example}

\noindent
Let $V_1 = [0, 100]$ and $V_2 = [0, 20]$ be the domains of two type gain criteria: $\mathcal{K}_1$ and $\mathcal{K}_2$, respectively. This means that $n = 2$, so $\mathit{CS}$ and $\mathit{US}$ will be $2$-dimensional. Additionally, let $\mathbf{w} = [1.0000, 0.5000]$, which implies $n_p = 2 = n$ (no zero weights) and means that also $\mathit{VS}$ will be $2$-dimensional.  

Because both criteria are of type gain, the re-scaling functions $\mathcal{U}_1(v)$ and $\mathcal{U}_2(v)$ corresponding to criteria $\mathcal{K}_1$ and $\mathcal{K}_2$, respectively, will be defined as follows:
\begin{itemize}
    \item $\mathcal{U}_1(v) = \frac{v-0}{100-0}$,
    \item $\mathcal{U}_2(v) = \frac{v-0}{20-0}$.
\end{itemize}
Consider an alternative represented as $E = [75, 10] \in \mathit{CS}$. Its representation in $\mathit{US}$ is found by an application of $\mathcal{U}_1(v)$ and $\mathcal{U}_2(v)$ to $E$ and turns out to
be 
{\small
\begin{align*}
\mathbf{u} &= [\mathcal{U}_1(75),\mathcal{U}_2(10)] = [\frac{75-0}{100-0},\frac{10-0}{20-0}] = \\
&= [0.7500, 0.5000].  
\end{align*}
}
Applying the weights vector $\mathbf{w}$ to $\mathbf{u}$ creates the $\mathit{VS}$ representation of $\mathbf{u}$ as 
{\small
\begin{align*}
\mathbf{v} &= \mathbf{w} \circ \mathbf{u} = [1.0000, 0.5000] \circ [0.7500, 0.5000] = \\
&= [0.7500, 0.2500].
\end{align*}
}

Now, projecting $\mathbf{v} = [0.7500, 0.2500]$ onto $\mathbf{w} = [1.0000, 0.5000]$ produces:
{\small
\begin{align*}
\mathbf{v} \!\searrow\! \mathbf{w} &= 
\frac{\mathbf{v}\cdot\mathbf{w}^T}{\norm{\mathbf{w}}_2^2} \mathbf{w} = \\
&= \frac{[0.7500, 0.2500]\cdot[1.0000, 0.5000]}{\norm{[1.0000, 0.5000]}_2^2} \cdot [1.0000, 0.5000] = \frac{0.7500 \cdot 1.0000 + 0.2500 \cdot 0.5000}{\sqrt{1.0000 \cdot 1.0000 + 0.5000 \cdot 0.5000}^2} \cdot [1.0000, 0.5000] = \\
&= \frac{0.7500 \cdot 1.0000 + 0.2500 \cdot 0.5000}{1.0000 \cdot 1.0000 + 0.5000 \cdot 0.5000} \cdot [1.0000, 0.5000] = 
\frac{0.875}{1.25} \cdot [1.0000, 0.5000] = 
0.7000 \cdot [1.0000, 0.5000] = \\
&= [0.7000, 0.3500].
\end{align*}
}
Simultaneously, rejecting $\mathbf{v} = [0.7500, 0.2500]$ from $\mathbf{w} = [1.0000, 0.5000]$ produces:
{\small
\begin{align*}
\mathbf{v} \!\nearrow\! \mathbf{w} &= 
\mathbf{v} - \mathbf{v} \!\searrow\! \mathbf{w} = \\
&= [0.7500, 0.2500] - [0.7500, 0.2500] \!\searrow\! [1.0000, 0.5000] =
[0.7500, 0.2500] - [0.7000, 0.3500] = \\
&= [0.0500, -0.1000].
\end{align*}
}
For the given $\mathbf{w} = [1.0000, 0.5000]$ one gets 
{\small
\begin{align*}
s = \frac{\norm{\mathbf{w}}_2}{mean(\mathbf{w})} = \frac{\norm{[1.0000, 0.5000]}_2}{mean([1.0000, 0.5000])} = \frac{1.1180}{0.7500} = 1.4907. 
\end{align*}
}
Applying this value in the definitions of $mean^{01}_{\mathbf{w}}(\mathbf{v})$ and $std^{01}_{\mathbf{w}}(\mathbf{v})$
results in:
{\small
\begin{align*}
mean^{01}_{\mathbf{w}}(\mathbf{v}) = \frac{\norm{\mathbf{v} \!\searrow\! \mathbf{w}}_2}{s} = \frac{\norm{[0.7000, 0.3500]}_2}{1.4907} = \frac{0.7826}{1.4907} = 0.5250,
\end{align*}
}
{\small
\begin{align*}
std^{01}_{\mathbf{w}}(\mathbf{v}) = \frac{\norm{\mathbf{v} \!\nearrow\! \mathbf{w}}_2}{s} = \frac{\norm{[0.0500, -0.1000]}_2}{s} = \frac{0.1118}{1.4907} = 0.0750.
\end{align*}
}

Because $\mathbf{\overline{v}} = \mathbf{v} \!\searrow\! \mathbf{w} = [0.7000, 0.3500]$ and $\mathbf{v} \!\nearrow\! \mathbf{w} = [0.0500, -0.1000]$, one gets:
{\small
\begin{align*}
\delta^{01}_{\mathbf{w}}(\mathbf{\overline{v}},\mathbf{0}) = \frac{\norm{\mathbf{\overline{v}} - \mathbf{0}}_2}{s} = \frac{\norm{[0.7000, 0.3500] - [0.0000, 0.0000]}_2}{s} = \frac{\norm{[0.7000, 0.3500]}_2}{s} = \frac{0.7826}{1.4907} = 0.5250
\end{align*}
}\quad \quad \, (clearly, $\delta^{01}_{\mathbf{w}}(\mathbf{\overline{v}},\mathbf{0}) = mean^{01}_{\mathbf{w}}(\mathbf{v}) = 0.5250$),
{\small
\begin{align*}
\delta^{01}_{\mathbf{w}}(\mathbf{\overline{v}},\mathbf{w}) = \frac{\norm{\mathbf{\overline{v}} - \mathbf{w}}_2}{s} = \frac{\norm{[0.7000, 0.3500] - [1.0000, 0.5000]}_2}{s} = \frac{\norm{[-0.3000, -0.1500]}_2}{s} = \frac{0.3354}{1.4907} = 0.2250
\end{align*}
}\quad \quad \, (clearly, $\delta^{01}_{\mathbf{w}}(\mathbf{\overline{v}},\mathbf{w}) = mean(\mathbf{w}) - mean^{01}_{\mathbf{w}}(\mathbf{v}) = 0.7500 - 0.5250 = 0.2250$),
{\small
\begin{align*}
\delta^{01}_{\mathbf{w}}(\mathbf{\overline{v}},\mathbf{v}) = \frac{\norm{\mathbf{\overline{v}} - \mathbf{v}}_2}{s} = \frac{\norm{[0.7000, 0.3500] - [0.7500, 0.2500]}_2}{s} = \frac{\norm{[-0.0500, -0.1000]}_2}{s} = \frac{0.1118}{1.4907} = 0.0750
\end{align*}
}\quad \quad \, (clearly, $\delta^{01}_{\mathbf{w}}(\mathbf{\overline{v}},\mathbf{v}) = std^{01}_{\mathbf{w}}(\mathbf{v}) = 0.0750$).
\vspace{8pt}

\begin{figure*}[h!]
\centering
\includegraphics[width=0.48\textwidth]  {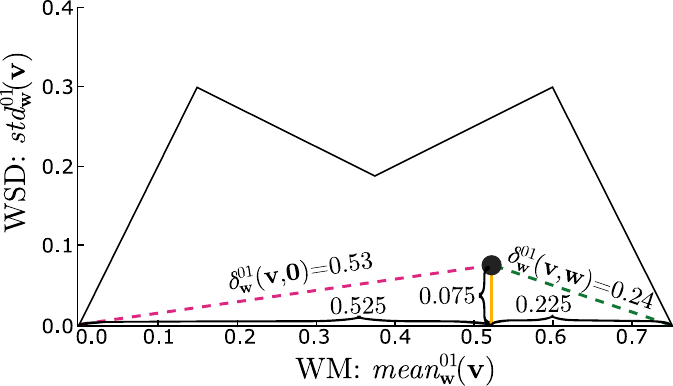}
\caption{An illustration of how the weight-scaled mean (WM) and standard deviation (WSD) define the WMSD-space by the IA-WMSD property.}
\label{fig:app:IA-WMSD}
\end{figure*}

Finalizing the example, we get:
{\small
\begin{align*}
\delta^{01}_{\mathbf{w}}(\mathbf{v},\mathbf{0}) = \frac{\norm{\mathbf{v}-\mathbf{0}}_2}{s} = \frac{\norm{[0.7500, 0.2500] - [0.0000, 0.0000]}_2}{1.4907} = \frac{\norm{[0.7500, 0.2500]}_2}{1.4907} = \frac{0.7906}{1.4907} = 0.5304
\end{align*}
}
{\small
\begin{align*}
\delta^{01}_{\mathbf{w}}(\mathbf{v},\mathbf{w}) = \frac{\norm{\mathbf{v}-\mathbf{w}}_2}{s} = \frac{\norm{[0.7500, 0.2500] - [1.0000, 0.5000]}_2}{1.4907} = \frac{\norm{[-0.2500, -0.2500]}_2}{1.4907} = \frac{0.3536}{1.4907} = 0.2372
\end{align*}
}
which allows to verify that: 
\begin{itemize}
\item $\sqrt{mean^{01}_\mathbf{w}(\mathbf{v})^2+std^{01}_\mathbf{w}(\mathbf{v})^2} = \sqrt{0.5250^2+0.0750^2} = \sqrt{0.2813} = 0.5304 = \delta^{01}_{\mathbf{w}}(\mathbf{v},\mathbf{0})$,
\item $\sqrt{mean(\mathbf{w}) - (mean^{01}_\mathbf{w}(\mathbf{v}))^2+std^{01}_\mathbf{w}(\mathbf{v})^2} = \sqrt{(0.7500 - 0.5250)^2+0.0750^2} = \sqrt{0.2250^2+0.0750^2} = \sqrt{0.0563} = 0.2372 = \delta^{01}_{\mathbf{w}}(\mathbf{v},\mathbf{w})$.
\end{itemize}
All these calculated values are shown in Figure~\ref{fig:app:IA-WMSD}.

What remains to be computed with $\delta^{01}_{\mathbf{w}}(\mathbf{v},\mathbf{0}) = 0.5304$ and $\delta^{01}_{\mathbf{w}}(\mathbf{v},\mathbf{w}) = 0.2327$ is:
\begin{itemize} 
\item $\mathsf{I}_{\mathbf{w}}(\mathbf{v}) = 1 - \frac{\delta^{01}_{\mathbf{w}}(\mathbf{v},\mathbf{w})}{mean(\mathbf{w})} = 1 - \frac{0.2327}{0.7500} = 1 - 0.3103 = 0.6897$,
\item $\mathsf{A}_{\mathbf{w}}(\mathbf{v}) = \frac{\delta^{01}_{\mathbf{w}}(\mathbf{v},\mathbf{0})}{mean(\mathbf{w})} = \frac{0.5304}{0.7500} = 0.7072$, 
\item $\mathsf{R}_{\mathbf{w}}(\mathbf{v}) = \frac{\delta^{01}_{\mathbf{w}}(\mathbf{v},\mathbf{0})}{\delta^{01}_{\mathbf{w}}(\mathbf{v},\mathbf{w})+\delta^{01}_{\mathbf{w}}(\mathbf{v},\mathbf{0})} = \frac{0.5304}{0.2327+0.5304} = 0.6951$.
\end{itemize}
All these aggregation values can also be observed color-coded in Figure~\ref{fig:WMSD-colors}.

\subsection{A $4$-dimensional example}

Let us now consider a fully analogous but slightly more-dimensional (precisely: $4$-dimensional) example to appreciate the actual dimensionality reduction that takes place in such a situation. In the previous example, a vector from $2$-dimensional $\mathit{CS}$ was transformed to equally $2$-dimensional WMSD-space. Now, a vector from $4$-dimensional $\mathit{CS}$ will be transformed to $2$-dimensional WMSD-space, although no information \textit{relevant to the TOPSIS method}  will be lost. This `lossless' character of the reduction concerns of course all $n > 2$.

Let $i \in \{1, 2, 3, 4\}$. Additionally, let $V_i = [0, 100]$ be the domains of criteria $\mathcal{K}_i$, all of type type gain.
This means that $n = 4$, so $\mathit{CS}$ and $\mathit{US}$ will be $4$-dimensional. Now, let $\mathbf{w} = [0.2500, 1.0000, 0.2500, 0.5000]$, which corresponds to the second sets of weights in the analysis of Index of Economic Freedom, as described in subsection~\ref{sec:freedom} of Section~\ref{sec:case}.
The vector of weights implies $n_p = 4 = n$ (no zero weights) and means that also $\mathit{VS}$ will be $4$-dimensional.  

Because all criteria are of type gain, the re-scaling functions $\mathcal{U}_i(v)$ corresponding to criteria $\mathcal{K}_i$ will be defined as follows: $\mathcal{U}_i(v) = \frac{v-0}{100-0}$.
Consider a vector $[62.43, 82.43, 75.37, 81.27] \in \mathit{CS}$, which is the representation of Chile. Its image in $\mathit{US}$ is found by an application of $\mathcal{U}_i(v)$ to $[62.43, 82.43, 75.37, 81.27]$ and turns out to be 
{\small
\begin{align*}
\mathbf{u} &= [\mathcal{U}_1(62.43),\mathcal{U}_2(82.43),\mathcal{U}_3(75.37),\mathcal{U}_4(81.27)] = [\frac{62.43-0}{100-0}, \frac{82.43-0}{100-0}, \frac{75.37-0}{100-0}, \frac{81.27-0}{100-0}] \\
&= [0.6243, 0.8243, 0.7537, 0.8127].  
\end{align*}
}
Applying the weights vector $\mathbf{w}$ to $\mathbf{u}$ creates the 
$\mathit{VS}$ representation of $\mathbf{u}$ as 
{\small
\begin{align*}
\mathbf{v} &= \mathbf{w} \circ \mathbf{u} = [0.2500, 1.0000, 0.2500, 0.5000] \circ [0.6243, 0.8243, 0.7537, 0.8127] = \\
&= [0.1561, 0.8243, 0.1884, 0.4063].
\end{align*}
}

Now, projecting $\mathbf{v} = [0.1561, 0.8243, 0.1884, 0.4063]$ onto $\mathbf{w} = [0.2500, 1.0000, 0.2500, 0.5000]$ produces: 
{\small 
\begin{align*}
\mathbf{v} \!\searrow\! \mathbf{w} &= \frac{\mathbf{v}\cdot\mathbf{w}^T}{\norm{\mathbf{w}}_2^2} \mathbf{w} = \\
&= \frac{[0.1561, 0.8243, 0.1884, 0.4063]\cdot[0.2500, 1.0000, 0.2500, 0.5000]}{\norm{[0.2500, 1.0000, 0.2500, 0.5000]}_2^2} \cdot [0.2500, 1.0000, 0.2500, 0.5000] = \\
&= \frac{0.1561 \cdot 0.25 + 0.8243 \cdot 1.00  + 0.1884 \cdot 0.25  + 0.4063 \cdot 0.50}{\sqrt{0.25 \cdot 0.25 + 1.00 \cdot 1.00  + 0.25 \cdot 0.25  + 0.50 \cdot 0.50}^2} \cdot [0.2500, 1.0000, 0.2500, 0.5000] = \\
&= \frac{0.1561 \cdot 0.25 + 0.8243 \cdot 1.00  + 0.1884 \cdot 0.25  + 0.4063 \cdot 0.50}{0.25 \cdot 0.25 + 1.00 \cdot 1.00  + 0.25 \cdot 0.25  + 0.50 \cdot 0.50} \cdot [0.2500, 1.0000, 0.2500, 0.5000] = \\
&= \frac{1.1136}{1.3750} \cdot [0.2500, 1.0000, 0.2500, 0.5000] = 
0.8099 \cdot [0.2500, 1.0000, 0.2500, 0.5000] = \\
&= [0.2025, 0.8099, 0.2025, 0.4049].
\end{align*}
}
Simultaneously, rejecting $\mathbf{v} = [0.1561, 0.8243, 0.1884, 0.4063]$ from $\mathbf{w} = [0.2500, 1.0000, 0.2500, 0.5000]$ produces:
{\small
\begin{align*}
\mathbf{v} \!\nearrow\! \mathbf{w} &= \mathbf{v} - \mathbf{v} \!\searrow\! \mathbf{w} = \\
&= [0.1561, 0.8243, 0.1884, 0.4063] - [0.1561, 0.8243, 0.1884, 0.4063] \!\searrow\! [0.2500, 1.0000, 0.2500, 0.5000] = \\
&= [0.1561, 0.8243, 0.1884, 0.4063] - [0.2025, 0.8099, 0.2025, 0.4049] = \\
&= [-0.0464,0.0144,-0.0141,0.0014].
\end{align*}
}

For the given $\mathbf{w} = [0.2500, 1.0000, 0.2500, 0.5000]$ one gets 
{\small
\begin{align*}
s = \frac{\norm{\mathbf{w}}_2}{mean(\mathbf{w})} = \frac{\norm{[0.2500, 1.0000, 0.2500, 0.5000]}_2}{mean([0.2500, 1.0000, 0.2500, 0.5000])} = \frac{1.1726}{0.5000} = 2.3452.
\end{align*}
}
Applying this value in the definitions of $mean^{01}_{\mathbf{w}}(\mathbf{v})$ and $std^{01}_{\mathbf{w}}(\mathbf{v})$
results in:
{\small
\begin{align*}
mean^{01}_{\mathbf{w}}(\mathbf{v}) &= \frac{\norm{\mathbf{v} \!\searrow\! \mathbf{w}}_2}{s} = \frac{\norm{[0.2025, 0.8099, 0.2025, 0.4049]}_2}{2.3452} = \frac{0.9497}{2.3452} = 0.4049,
\end{align*}
}
{\small
\begin{align*}
std^{01}_{\mathbf{w}}(\mathbf{v}) &= \frac{\norm{\mathbf{v} \!\nearrow\! \mathbf{w}}_2}{s} = \frac{\norm{[-0.0464,0.0144,-0.0141,0.0014]}_2}{s} = \frac{0.0506}{2.3452} = 0.0216.
\end{align*}
}

The above coordinates of Chile (CHL) in WMSD-space can noticed on Figure~\ref{fig:freedom}B.

Because $\mathbf{\overline{v}} = \mathbf{v} \!\searrow\! \mathbf{w} = [0.2025, 0.8099, 0.2025, 0.4049]$ and $\mathbf{v} \!\nearrow\! \mathbf{w} = [-0.0464,0.0144,-0.0141,0.0014]$, one gets:
{\small
\begin{align*}
\delta^{01}_{\mathbf{w}}(\mathbf{\overline{v}},\mathbf{0}) &= \frac{\norm{\mathbf{\overline{v}} - \mathbf{0}}_2}{s} = \\
&= \frac{\norm{[0.2025, 0.8099, 0.2025, 0.4049] - [0.000, 0.000, 0.000, 0.000]}_2}{s} = \frac{\norm{[0.2025, 0.8099, 0.2025, 0.4049]}_2}{s} = \\
&= \frac{0.9497}{2.3452} = 0.4049
\end{align*}
}\quad \quad \, (clearly, $\delta^{01}_{\mathbf{w}}(\mathbf{\overline{v}},\mathbf{0}) = mean^{01}_{\mathbf{w}}(\mathbf{v}) = 0.4049$),
{\small
\begin{align*}
\delta^{01}_{\mathbf{w}}(\mathbf{\overline{v}},\mathbf{w}) &= \frac{\norm{\mathbf{\overline{v}} - \mathbf{w}}_2}{s} = \\
&= \frac{\norm{[0.2025, 0.8099, 0.2025, 0.4049] - [0.2500, 1.0000, 0.2500, 0.5000]}_2}{s} = \frac{\norm{-0.0475,-0.1901,-0.0475,-0.0951}_2}{s} = \\
&= \frac{0.2229}{2.3452} = 0.0950
\end{align*}
}\quad \quad \, (clearly, $\delta^{01}_{\mathbf{w}}(\mathbf{\overline{v}},\mathbf{w}) = mean(\mathbf{w}) - mean^{01}_{\mathbf{w}}(\mathbf{v}) = 0.5000 - 0.4049 = 0.0950$),
{\small
\begin{align*}
\delta^{01}_{\mathbf{w}}(\mathbf{\overline{v}},\mathbf{v}) &= \frac{\norm{\mathbf{\overline{v}} - \mathbf{v}}_2}{s} = \\
&= \frac{\norm{[0.2025, 0.8099, 0.2025, 0.4049] - [0.1561, 0.8243, 0.1884, 0.4063]}_2}{s} = \frac{\norm{[0.0464,-0.0144,0.0141,-0.0014]}_2}{s} = \\
&= \frac{0.0506}{2.3452} = 0.0216
\end{align*}
}\quad \quad \, (clearly, $\delta^{01}_{\mathbf{w}}(\mathbf{\overline{v}},\mathbf{v}) = std^{01}_{\mathbf{w}}(\mathbf{v}) = 0.0216$).
\vspace{8pt}

Finalizing the example, we get:
{\small
\begin{align*}
\delta^{01}_{\mathbf{w}}(\mathbf{v},\mathbf{0}) &= 
\frac{\norm{\mathbf{v}-\mathbf{0}}_2}{s} = \\
&= \frac{\norm{[0.1561, 0.8243, 0.1884, 0.4063] - [0.000, 0.000, 0.000, 0.000]}_2}{2.3452} = \frac{\norm{[0.1561, 0.8243, 0.1884, 0.4063]}_2}{2.3452} = \\
&= \frac{0.9510}{2.3452} = 0.4055
\end{align*}
}
{\small
\begin{align*}
\delta^{01}_{\mathbf{w}}(\mathbf{v},\mathbf{w}) &= 
\frac{\norm{\mathbf{v}-\mathbf{w}}_2}{s} = \\
&= \frac{\norm{[0.1561, 0.8243, 0.1884, 0.4063] - [0.2500, 1.0000, 0.2500, 0.5000]}_2}{2.3452} = \frac{\norm{[-0.0939,-0.1757,-0.0616,-0.0937]}_2}{2.3452} = \\
&= \frac{0.2286}{2.3452} = 0.0975
\end{align*}
}

which allows us to verify that: 
\begin{itemize}
\item $\sqrt{mean^{01}_\mathbf{w}(\mathbf{v})^2+std^{01}_\mathbf{w}(\mathbf{v})^2} = \sqrt{0.4049^2+0.0216^2} = \sqrt{0.1644} = 0.4054 = \delta^{01}_{\mathbf{w}}(\mathbf{v},\mathbf{0})$,
\item $\sqrt{mean(\mathbf{w}) - (mean^{01}_\mathbf{w}(\mathbf{v}))^2+std^{01}_\mathbf{w}(\mathbf{v})^2} = \sqrt{(0.5000 - 0.4049)^2+0.0216^2} = \sqrt{0.0951^2+0.0216^2} = \\\sqrt{0.0095} = 0.0975 = \delta^{01}_{\mathbf{w}}(\mathbf{v},\mathbf{w})$.
\end{itemize}
What remains to be computed with $\delta^{01}_{\mathbf{w}}(\mathbf{v},\mathbf{0}) = 0.4055$ and $\delta^{01}_{\mathbf{w}}(\mathbf{v},\mathbf{w}) = 0.0975$ is:
\begin{itemize} 
\item $\mathsf{I}_{\mathbf{w}}(\mathbf{v}) = 1 - \frac{\delta^{01}_{\mathbf{w}}(\mathbf{v},\mathbf{w})}{mean(\mathbf{w})} = 1 - \frac{0.0975}{0.5000} = 1 - 0.1949 = 0.8051$,
\item $\mathsf{A}_{\mathbf{w}}(\mathbf{v}) = \frac{\delta^{01}_{\mathbf{w}}(\mathbf{v},\mathbf{0})}{mean(\mathbf{w})} = \frac{0.5055}{0.5000} = 0.8110$, \item $\mathsf{R}_{\mathbf{w}}(\mathbf{v}) = \frac{\delta^{01}_{\mathbf{w}}(\mathbf{v},\mathbf{0})}{\delta^{01}_{\mathbf{w}}(\mathbf{v},\mathbf{w})+\delta^{01}_{\mathbf{w}}(\mathbf{v},\mathbf{0})} = \frac{0.4055}{0.0975+0.4055} = 0.8062$.
\end{itemize}
This final value ($0.8062$) can also be seen in column 4 of the first row of Table~\ref{tab:SA_R_rankings} (three significant digits).

\section{The Lower and Upper Boundary of the WMSD-space}
\label{app:perimeters_of_WMSD-space}
\noindent
Recall that 
$mean^{01}_{\mathbf{w}}(\mathbf{v})$ is defined as the re-scaled length of a projection of $\mathbf{v} \in \mathit{VS}$ onto $\mathbf{w}$, 
$std^{01}_{\mathbf{w}}(\mathbf{v})$ is defined as the re-scaled length of a rejection of $\mathbf{v} \in \mathit{VS}$ from $\mathbf{w}$,
and that the diagonal $D_{\mathbf{0}}^{\mathbf{w}}$ equals $\{d \cdot \mathbf{w} \; | \; d \in [0, 1]\}$.

Additionally, given $\mathbf{w}$ such that $n_p \geq 2$, and $t \in [0, mean(\mathbf{w})]$, consider the following parametrized (non-linear) programming problems:

\vspace{-10pt}
\begin{equation*}
\begin{array}{ll}
\text{find} \; p_l(t) \; \text{as:} & \displaystyle\min\limits_{\mathbf{v} \in VS} std^{01}_{\mathbf{w}}(\mathbf{v})\\
\text{subject to:} &\displaystyle mean^{01}_{\mathbf{w}}(\mathbf{v}) = t\\
\end{array}
\end{equation*}

and 

\vspace{-10pt}
\begin{equation*}
\begin{array}{ll}
\text{find} \; p_u(t) \; \text{as:} & \displaystyle\max\limits_{\mathbf{v} \in VS} std^{01}_{\mathbf{w}}(\mathbf{v})\\
\text{subject to:} &\displaystyle mean^{01}_{\mathbf{w}}(\mathbf{v}) = t\\
\end{array}
\end{equation*}

\noindent
where $t$ is the parameter.

For all $t \in [0, mean(\mathbf{w})]$ the resulting vectors: 
\begin{itemize}
\item $[t,p_l(t)]$ represent points of the `lower perimeter' of WMSD-space,
\item $[t,p_u(t)]$ represent points of the `upper perimeter' of WMSD-space.
\end{itemize}

\noindent
As $p_l(t) = 0$ for all $t \in [0, mean(\mathbf{w})]$, the `lower perimeter' of WMSD-space simply constitutes a horizontal segment. 
This is because in $\mathit{VS}$ the re-scaled length of a rejection of $\mathbf{v} \in \mathit{VS}$ from $\mathbf{w}$ (i.e. the `orthogonally' computed distance between $\mathbf{v}$ and $D_{\mathbf{0}}^{\mathbf{w}}$), is minimal when $\mathbf{v} \in D_{\mathbf{0}}^{\mathbf{w}}$ and equals $0$ in such cases.

On the other hand, $p_u(t) = 0$ for $t \in \{0, mean(\mathbf{w})\}$, while $p_u(t) > 0$ for $t \in (0, mean(\mathbf{w}))$ (it also satisfies the following symmetry condition: $p_u(\frac{mean(\mathbf{w})}{2}-h) = p_u(\frac{mean(\mathbf{w})}{2}+h)$ for every $h \in [0, \frac{mean(\mathbf{w})}{2}]$).
This is because in $\mathit{VS}$ the re-scaled length of a rejection of $\mathbf{v} \in \mathit{VS}$ from $\mathbf{w}$ (i.e. the `orthogonally' computed distance between $\mathbf{v}$ and $D_{\mathbf{0}}^{\mathbf{w}}$) is $0$ when $\mathbf{v} \in \{\mathbf{0}, \mathbf{w}\}$ and greater than $0$ otherwise. In these cases, the maxima are attained at the borders of $\mathit{VS}$ and, because of its intricate shape, are dependent of $mean^{01}_{\mathbf{w}}(\mathbf{v})$. 
Unfortunately, the closed-form formula for these maxima, and thus for the `upper perimeter' of WMSD-space is unknown in the general case (as opposed to MSD-space~\cite{MSD-space}, which constitutes a special case of WMSD-space for $\mathbf{w} = \mathbf{1}$).



\end{document}